\PassOptionsToPackage{numbers,sort&compress}{natbib}
\PassOptionsToPackage{table,dvipsnames}{xcolor}
\PassOptionsToPackage{section}{placeins}
\documentclass[]{style/company_light}
\usepackage{hyperref}
\usepackage{url}

\usepackage[T1]{fontenc}
\usepackage[utf8]{inputenc}
\usepackage{seqsplit}
\usepackage[dvipsnames]{xcolor}
\usepackage{booktabs}
\usepackage{longtable}
\usepackage{amsfonts}
\usepackage{amsmath}
\usepackage{amssymb}
\usepackage{nicefrac}
\usepackage{microtype}
\usepackage{graphicx}
\usepackage{float}
\usepackage{placeins}
\usepackage{caption}
\usepackage{subcaption}
\usepackage{multirow}
\usepackage{enumerate}
\usepackage{enumitem}
\usepackage{adjustbox}
\usepackage{array}
\usepackage[skins]{tcolorbox}

\titleformat*{\paragraph}{\sffamily\bfseries}
\hypersetup{
  pdftitle={PPTBench: Can Coding Agents Reconstruct the Visual World through Structured, Editable Slides?},
  pdfauthor={Xiaoqiu Wang, Yizhe Chi, Wenyi Li, Deyao Hong, Zhihan Shan, Mingju Gao, Kaisen Yang, Youjie Zheng, Calvin Xiao, Qinhuai Na}
}

\NewDocumentCommand{\todo}
{ mO{} }{\textcolor{magenta}{\textsuperscript{\textit{TODO}}\textsf{\textbf{\small[#1]}}}}

\title{PPTBench: Can Coding Agents Reconstruct the Visual World through Structured, Editable Slides?}

\author{
  \footnotesize Xiaoqiu Wang$^{*}$, Yizhe Chi$^{*}$, Wenyi Li$^{*}$, Deyao Hong, Zhihan Shan, Mingju Gao, Kaisen Yang, Youjie Zheng, Calvin Xiao, Qinhuai Na$^{\ddagger}$
}
\renewcommand\affiliationformat[2][]{\makebox[\linewidth][c]{\small\bfseries #2}}
\affiliation{Navers Lab, Einsia.AI\quad Tsinghua University}
\renewcommand\contributionformat[2][]{%
  \vskip 0.15cm
  \makebox[\linewidth][c]{\footnotesize\color{gray}#2}%
}
\contribution{$^{*}$Equal Contribution\quad$^{\ddagger}$Corresponding Author}

\abstract{%
Coding agents are increasingly moving beyond text-based software tasks to reconstruct visual targets through code. This capability, visual coding, requires agents to translate their understanding of visual targets into executable code. Slides provide a natural testbed for this capability, combining rich visual structure with objects that can be programmatically created and edited. To measure this capability, we introduce \textbf{PPTBench}, a benchmark for reconstructing scientific flow diagrams as editable PowerPoint slides.
PPTBench covers 500 scientific flow-diagram tasks across 10 presentation domains, drawn from real research papers, and is evaluated with a four-stage agentic judge covering artifact validity, process and connector fidelity, rendering quality, and visual fidelity. Evaluation of ten models across 36 model--harness--effort configurations shows a substantial gap in reliable visual coding: the best configuration achieves 77.34, while the median across configurations is 24.38. Fine-grained analysis shows that agents can generally produce valid slide files, but struggle to produce high-quality reconstructions that faithfully recover the semantics and visual structure of the target. Further analysis shows that increasing reasoning effort primarily improves hard-gate passage rather than mean detail quality on each configuration's passing tasks, while configurations with more inspection tend to achieve higher overall scores. PPTBench establishes a measurable testbed for studying visual coding and advancing agents toward more reliable visual creation.
}

\date{September 28, 2026}
\metadata[Homepage]{\url{https://lab.einsia.ai/pptbench}}
\correspondence{\email{nana@einsia.ai}}

\begin{document}

\maketitle

\vspace{0.10in}
\begin{figure}[H]
  \centering
  \includegraphics[width=1.0\textwidth]{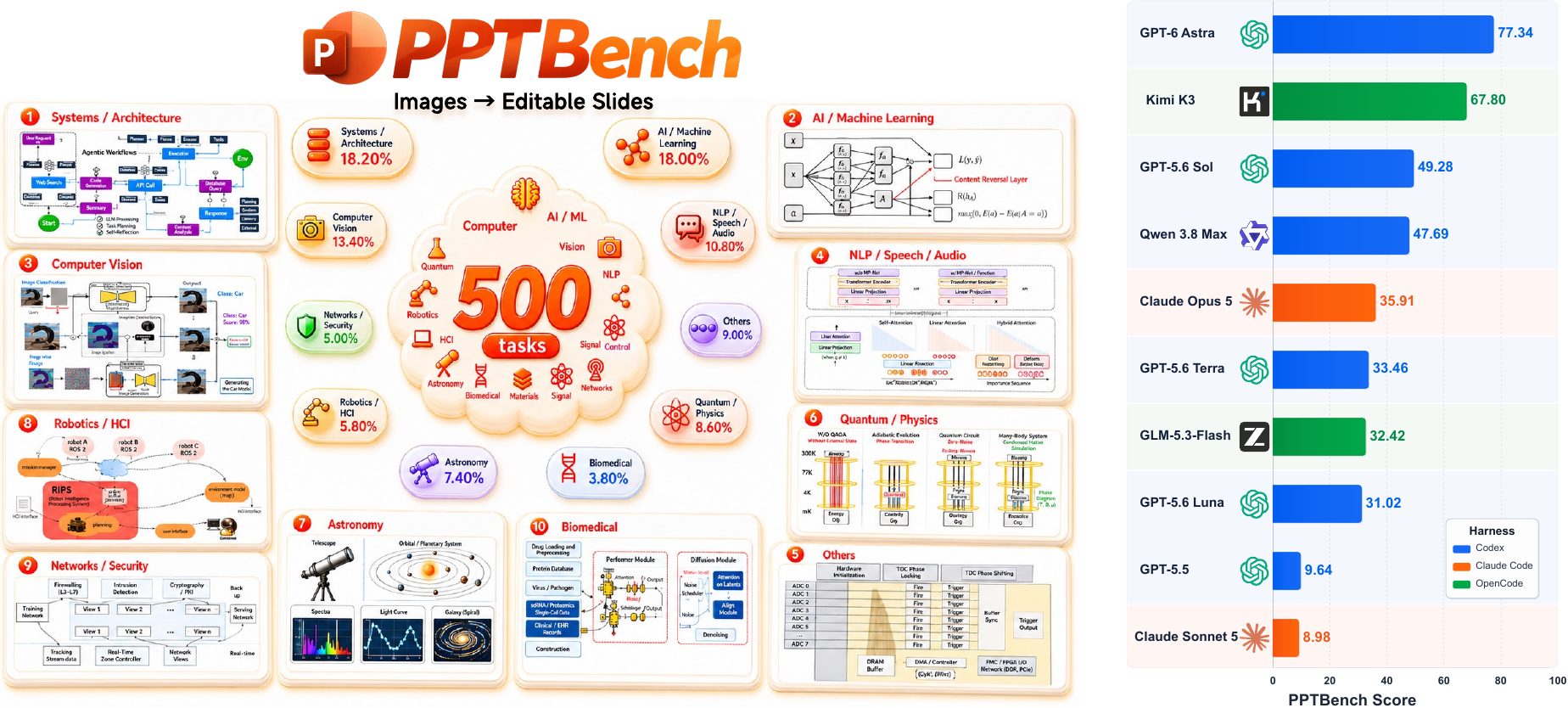}
  \caption{PPTBench task domains and representative model scores on 500 scientific flow diagrams.}
  \label{fig:teaser}
\end{figure}

\clearpage

\section{Introduction}
\label{sec:introduction}

Coding agents are beginning to act in the visual world. 
Instead of producing software evaluated mainly through APIs or terminals, they are increasingly asked to create webpages, interfaces, games, 3D scenes, diagrams, and documents~\citep{li2025sketch2code,chi2026gamedevbench,hu2024scenecraft,belouadi2024detikzify,ge2025autopresent}, requiring them to think through code as a medium for expressing visual ideas.
These tasks require more than generating valid programs: agents must infer the visual structure of a target and faithfully represent it in an executable form. This creates a fundamental challenge in aligning visual understanding with programmatic representation.

Slides provide an ideal testbed for the visual coding task above. A slide is not simply an image, but an editable composition whose elements have distinct types and spatial relationships. Its shapes, text, diagrams, and layout form a compact visual language for organizing and communicating information. Slides are also a pervasive medium for human knowledge: people use them to explain ideas, present research, teach concepts, and communicate decisions. Reconstructing a slide therefore goes beyond reproducing its appearance; it requires an agent to recover how visual elements are organized to convey meaning and express that structure in an editable artifact.

Existing benchmarks evaluate different parts of this capability. Slide-generation suites \citep{zheng2025pptagent,yang2026slidesgenbench,chen2026presentbench} start from documents or requests and evaluate open-ended designs. Chart-to-code and screenshot-to-markup benchmarks \citep{yang2025chartmimic,wu2024plot2code,tang2025chart2code,zhang2026realchart2code,si2024design2code,laurencon2024websight,li2025sketch2code} reconstruct fixed targets as code or markup. Office benchmarks such as PPTC~\citep{guo2023pptc} evaluate instructed edits to PPTX files, while Slide2Code~\citep{tang2025slidecoder} directly evaluates image-to-editable-slide reconstruction. The earlier benchmark named PPTBench~\citep{huang2025pptbench} spans slide detection, understanding, modification, and generation, including generation from screenshots. Our PPTBench focuses on scientific process diagrams and tests whether native editable reconstructions preserve their process and connector structure. Its evaluation gates artifact validity, process fidelity, and legibility before scoring visual detail, separating structural failures from appearance errors.

To rigorously measure and advance this capability, we introduce \textbf{PPTBench}, a benchmark for visual coding through editable slide reconstruction. Its 500 tasks are drawn from scientific flow diagrams in real arXiv papers, with each task requiring an agent to reconstruct a given diagram as a single-page PowerPoint presentation composed of native, editable objects.
The benchmark combines automated checks with human screening to ensure that the tasks are visually meaningful and structurally suitable for reconstruction.
To enable automated evaluation, PPTBench employs a four-stage Agentic Judge. It first checks file validity and native editable content, screening raster images for copies of the reference. It then evaluates whether the reconstructed objects preserve the semantics of the target, whether the rendered slide is visually and textually legible, and whether individual elements match the target in their geometry and textual content.
The four stages form a gated evaluation, so fundamental failures cannot be hidden by strong performance on other aspects. The score is computed from structured findings, and blinded validation shows substantial agreement with human judgments.

We evaluate ten models across 36 model--harness--effort configurations on all 500 tasks, yielding 18{,}000 scored reconstructions. The best configuration, GPT-6 Astra with high reasoning effort, reaches 77.34 out of 100, while the configuration median is 24.38, revealing a substantial gap in reliable visual coding. Agents usually produce usable slides, but only 31.74\% of reconstructions pass all gates, with semantic recovery as the dominant bottleneck. Text accounts for 53.0\% of detail deductions, with unintended wrapping as the largest source. We further find that reasoning effort mainly improves hard-gate pass rates, while inspection frequency is more strongly associated with higher scores than revision frequency. These findings distinguish file validity from faithful reconstruction of a diagram's process and visual structure.

In summary, the primary contributions of this paper are:

\begin{enumerate}[nosep]
  \item We identify the core challenge of visual coding as recovering visual structure in executable representations, and choose editable slides as a suitable medium for measuring this capability.

  \item We introduce PPTBench with 500 diverse scientific flow-diagram reconstruction tasks drawn from real arXiv papers, together with an automated four-stage Agentic Judge that verifies artifact validity, semantic correctness, rendering quality, and fine-grained visual fidelity.

  \item Through extensive experiments on frontier models, we uncover systematic limitations of current agents, such as gaps in semantic recovery and visual fidelity, and provide directions for advancing visual coding agents.
\end{enumerate}



\section{The PPTBench Benchmark}
\label{sec:benchmark}

PPTBench asks whether an agent can reconstruct a scientific flow diagram as a native, editable
PPTX slide from a single image. We first define the task and artifact contract, then describe the benchmark construction and the four-stage evaluation protocol.

\subsection{Task formulation}
\label{subsec:task_formulation}

A diagram reconstruction task provides an agent with a reference image $I$ and requires it to
generate a single-slide PPTX artifact $P=A(I)$. The reference image $I$ is a rasterized rendering
of a flow diagram from a real research paper and serves as the agent's only observation of the
target. The generated artifact must recover both the visual appearance and structural semantics
of the diagram while remaining natively editable. Specifically, the output should be composed of
native PowerPoint objects, except for explicitly approved raster elements that cannot be faithfully
represented with native primitives.

The native-object requirement prevents trivial solutions such as directly embedding the reference
image into the slide, ensuring that the task evaluates structural reconstruction rather than image
copying. The evaluator renders the generated PPTX with a fixed rendering environment, producing a
rendered image $\hat{I}=R(P)$, and compares it with the reference image $I$.

\subsection{Benchmark construction and coverage}
\label{subsec:construction}

\begin{figure}[t]
  \centering
  \includegraphics[width=\textwidth]{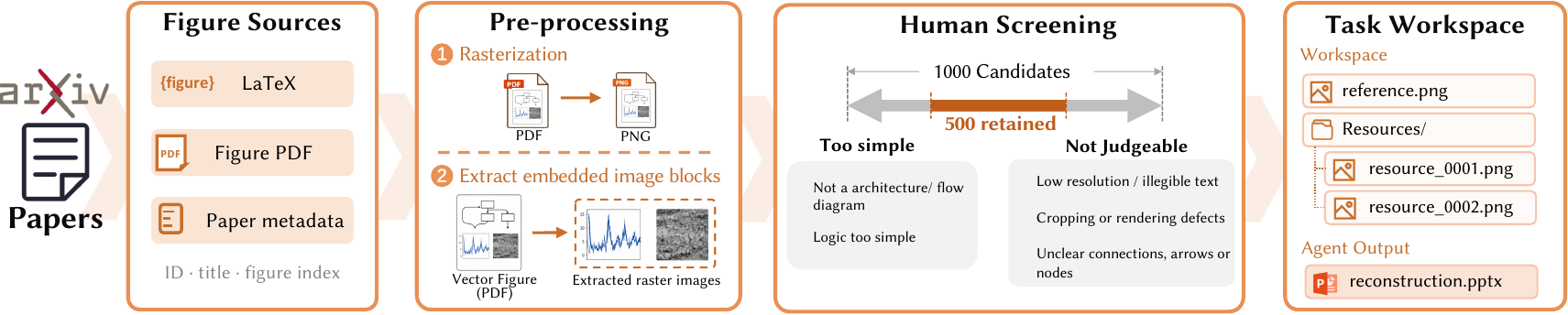}
  \caption{PPTBench construction: source retrieval, preprocessing, human screening, and task assembly. Screening retains 500 of 1{,}000 candidate diagrams.}
  \label{fig:pipeline}
\end{figure}

This section describes where candidate diagrams come from, how they are prepared and screened into a fixed set of reconstruction tasks, and which scientific domains the PPTBench covers. Figure~\ref{fig:pipeline} summarizes the construction process, from source retrieval and preprocessing to human screening and task assembly.

\paragraph{Figure sources.} We query arXiv metadata and download source archives,
then parse LaTeX \texttt{figure} environments to locate the authors' original graphics files.
Captions, file paths, and asset properties help us identify flow, system-architecture, and module
diagrams. Using published figures exposes agents to the irregular layouts, mixed notation, and
visual conventions that researchers use to explain their work. We retain each candidate's paper
identifier, title, version, figure number, and related metadata so that every task can be traced
back to its source.

\paragraph{Preprocessing.} We render each candidate directly from its original asset and record
the source identity and rendering parameters for reproducibility. Embedded raster blocks are
inspected separately. Self-contained elements that cannot reasonably be recreated with native
PowerPoint primitives may be approved for reuse, while the rest of the diagram must be rebuilt
as editable objects. We then remove repeated figures and alternate exports within each paper.
Filtering and preprocessing leave 1{,}000 candidates for human screening
(Figure~\ref{fig:pipeline}).

\paragraph{Human screening.} Annotators review the 1{,}000 candidates and retain 500 using five
exclusion criteria. We remove figures outside the flow and architecture categories and diagrams
whose content and logic are too simple to distinguish model capabilities. We also exclude
references with insufficient resolution or illegible text, as well as those with cropping errors,
source corruption, or rendering defects. Finally, annotators must be able to consistently identify
the key connections, arrow directions, and node boundaries. If the reference itself leaves an
arrow's direction unclear, a reconstruction cannot be judged reliably against it. Together, these
criteria select diagrams that are challenging enough to reveal differences between systems and
clear enough to evaluate. The resulting 500 tasks form the fixed evaluation set.

\paragraph{Task inputs.} Each task provides a reference image and, where needed, manually
approved raster assets. Agents inspect and measure the reference, then produce a single-slide
reconstruction with editable text, shapes, and connectors. Approved raster assets must appear
unchanged in their intended locations. The agent receives task-specific inputs without source-paper
provenance or access to other tasks.

\paragraph{Benchmark composition and coverage.}
\label{subsec:composition}
PPTBench draws from multiple scientific fields to evaluate reconstruction across
different visual structures and conventions. Its 500 tasks span 50 arXiv primary
categories, grouped into 10 representative domains in Figure~\ref{fig:teaser}. 
For example, systems workflows use branches and
feedback loops whose connections determine the process, while neural
architectures combine repeated modules with dense mathematical labels.
Quantum circuits introduce a different spatial convention, where gate
placement along aligned wires encodes the order of operations. These
examples require agents to recover both the visual arrangement and the
relationships it expresses across different diagram conventions.
Detailed category counts and source examples are provided in
Appendix~\ref{app:task_catalog}.

\subsection{Evaluation protocol}
\label{subsec:protocol}

\begin{figure}[t]
  \centering
  \includegraphics[width=\textwidth]{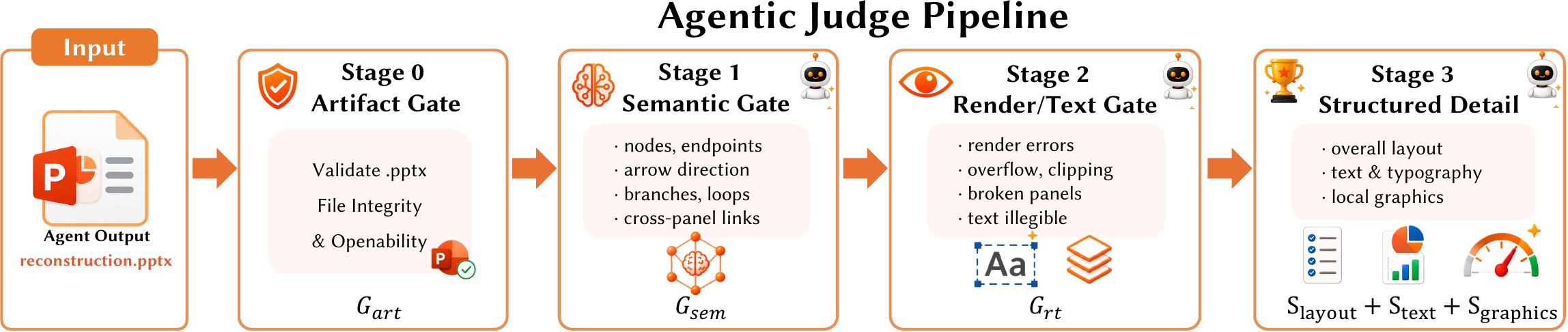}
  \caption{The four-stage Agentic Judge: artifact validation ($G_{\mathrm{art}}$), semantic
  correctness ($G_{\mathrm{sem}}$), rendering and text quality ($G_{\mathrm{rt}}$), and visual
  detail. Stages~1--3 run in three independent rounds. At least two rounds must pass both
  judged gates before merged detail findings determine the score (Eq.~\eqref{eq:score}).}
  \label{fig:protocol}
\end{figure}

Each reconstruction receives a score from 0 to 100, combining three validity gates with visual
detail quality (Figure~\ref{fig:protocol}):
\begin{equation}
  S =
  \underbrace{G_{\mathrm{art}}}_{\text{Stage 0}}
  \cdot
  \underbrace{G_{\mathrm{sem}}}_{\text{Stage 1}}
  \cdot
  \underbrace{G_{\mathrm{rt}}}_{\text{Stage 2}}
  \cdot
  \underbrace{\left(S_{\mathrm{layout}} + S_{\mathrm{text}} + S_{\mathrm{graphics}}\right)}_{\text{Stage 3}}.
  \label{eq:score}
\end{equation}
Artifact validation is deterministic, and GPT-5.6 Luna High evaluates Stages~1--3 in three
independent rounds. The binary gate $G_{\mathrm{art}}$ records artifact validity.
$G_{\mathrm{sem}}=1$ requires at least two rounds to pass Stage~1; $G_{\mathrm{rt}}=1$ requires
at least two of those rounds to also pass Stage~2. Each gate is zero otherwise, so a candidate
must pass both judged gates in at least two rounds to receive a detail score. Findings from
passing rounds are merged and converted into scores by deterministic rules. Layout, text, and
local graphics contribute up to 30, 40, and 30 points, respectively, with normalized detail
quality $Q=(S_{\mathrm{layout}}+S_{\mathrm{text}}+S_{\mathrm{graphics}})/100$.

The judge can write cropping, color-segmentation, and geometric-measurement scripts to inspect
the reference and reconstruction. Visual inspection checks the depicted process even when
equivalent diagrams use different PowerPoint object structures.
Section~\ref{subsec:validation_main} evaluates gate decisions against blinded human labels and
independent repetitions.

\paragraph{Stage 0: artifact validity.} The evaluator checks that the PPTX can be parsed and
rendered, contains exactly one slide and at least three native editable components, and has
no external relationships, embedded fonts, attachments, or ActiveX payloads. Missing or invalid
artifacts receive zero. The validator checks object bounds and screens raster images for copies
of the reference. Valid slides are rendered and paired with the
reference after cropping outer whitespace and normalizing scale and centering. Full-resolution
images remain available for inspection; Appendix~\ref{subsec:package} gives the image preparation
details.

\paragraph{Stage 1: semantic correctness.} The judge compares nodes, labels, and directed
connections, checking for missing, added, or incorrectly merged steps, wrong endpoints or
directions, altered branches, merges, loops, or cross-panel links, and label errors that change
meaning. It also traces local connector segments and
intermediate arrowheads, so a missing segment or moved arrowhead can fail this stage even when
the overall process looks plausible. A confirmed discrepancy ends the round, with the affected
node or connection recorded as the reason.

\paragraph{Stage 2: rendering and text quality.} After the semantic gate passes, the judge
checks legibility. This gate triggers when black regions, distortion, tearing, severe ghosting,
or blank blocks affect more than 30\% of the figure, or when more than 50\% of text boxes
overflow, overlap, or break improperly. A failure ends the round. Localized defects proceed to
Stage~3 for graded deductions.

\paragraph{Stage 3: fine-grained visual quality.} The judge compares layout and composition,
text and typography, and local graphics such as nodes and connectors. It records defects and
their extent without assigning points. Findings from passing rounds are merged by issue type
within each category and dimension, retaining the largest reported affected fraction.
A fixed rubric then deducts points according to severity and extent, up to each dimension's
score cap. Outer whitespace, global centering, and absolute content scale do not affect the
score. Appendix~\ref{app:detail_taxonomy} gives the finding schema, merging rules, and deduction
formulas, and Appendix~\ref{app:case_study} provides scored examples.

\section{Diagnosing Visual Coding Capabilities on PPTBench}
\label{sec:experiments}

\subsection{Setup and Metrics}
\label{subsec:setup_main}

We evaluate ten models across 36 model--harness--effort configurations
(Table~\ref{tab:main_leaderboard}). Each configuration runs all 500 tasks under the task and
evaluation protocol in Section~\ref{sec:benchmark}, yielding 18{,}000 scored reconstructions.
Models use Codex, Claude Code, or OpenCode as listed in the table, so results reflect both the
model and its harness. The evaluation covers the full benchmark, including the tasks used
for protocol development and judge validation.

Our main metric is mean $S$ over all 500 tasks, with invalid or gated outputs scored as zero.
Artifact validity and gate-pass rate measure the fractions passing Stage~0 and all gates,
respectively. Score on passed tasks is the mean score computed only over
reconstructions that pass all gates; multiplying it by the gate-pass fraction gives mean $S$.
Layout, text, and local-graphics scores also
average over all tasks, with zeros for gated outputs, and sum to mean $S$. We report generation
cost in USD for each configuration's 500-task evaluation, excluding judging.
Appendix~\ref{subsec:setup} gives execution and cost-accounting details;
Appendix~\ref{app:paired_analysis} compares configuration and model-family weighting.

\subsection{Overall Performance}
\label{subsec:main_results_main}

\textbf{The best configuration scores 77.34, while the median scores 24.38.}
Table~\ref{tab:main_leaderboard} reports all 36 model--harness--effort configurations, ranked by mean score. GPT-6 Astra at high leads with 77.34, followed by Astra xhigh (72.60) and Astra max (68.76). Eight of 36 configurations exceed 40 points, and the weakest, GPT-5.6 Luna at none, scores 1.53. Across all 18{,}000 model--task pairs, 461 (2.56\%) pass the gates without any recorded detail finding.

\providecommand{\lbicon}[1]{\raisebox{-0.12ex}{\includegraphics[height=1.05em]{assets/model_icons/#1.png}}}
\providecommand{\lbup}{\raisebox{0.15ex}{\scriptsize$\uparrow$}}
\providecommand{\lbdn}{\raisebox{0.15ex}{\scriptsize$\downarrow$}}

\begin{table}[!t]
  \caption{Overall performance of frontier agents on PPTBench across 36 model--harness--effort configurations, ranked by mean score over 500 reconstruction tasks.}
  \label{tab:leaderboard}
  \label{tab:main_leaderboard}
  \centering
  \scriptsize
  \setlength{\tabcolsep}{5pt}
  \renewcommand{\arraystretch}{1.08}
  \newcommand{\lbhead}[1]{\begin{tabular}[c]{@{}c@{}}#1\end{tabular}}
  \begin{adjustbox}{max width=\linewidth}
  \begin{tabular}{@{}lll @{\hspace{10pt}} r @{\hspace{10pt}} rr @{\hspace{10pt}} rrr @{\hspace{10pt}} r@{}}
    \toprule
    \textbf{Model} & \textbf{Harness} & \textbf{Effort}
      & \multicolumn{1}{c}{\textbf{Mean score}\,\lbup}
      & \multicolumn{1}{c}{\lbhead{\textbf{Gate pass}\\\textbf{rate}\,\lbup}}
      & \multicolumn{1}{c}{\lbhead{\textbf{Score on}\\\textbf{passed tasks}\,\lbup}}
      & \multicolumn{1}{c}{\textbf{Layout}}
      & \multicolumn{1}{c}{\textbf{Text}}
      & \multicolumn{1}{c}{\textbf{Graphics}}
      & \multicolumn{1}{c@{}}{\textbf{Cost}\,\lbdn} \\
    & & & \multicolumn{1}{c}{/100} & \multicolumn{1}{c}{(\%)}
      & \multicolumn{1}{c}{/100} & \multicolumn{1}{c}{/30}
      & \multicolumn{1}{c}{/40} & \multicolumn{1}{c}{/30}
      & \multicolumn{1}{c@{}}{(\$)} \\
    \midrule
    \lbicon{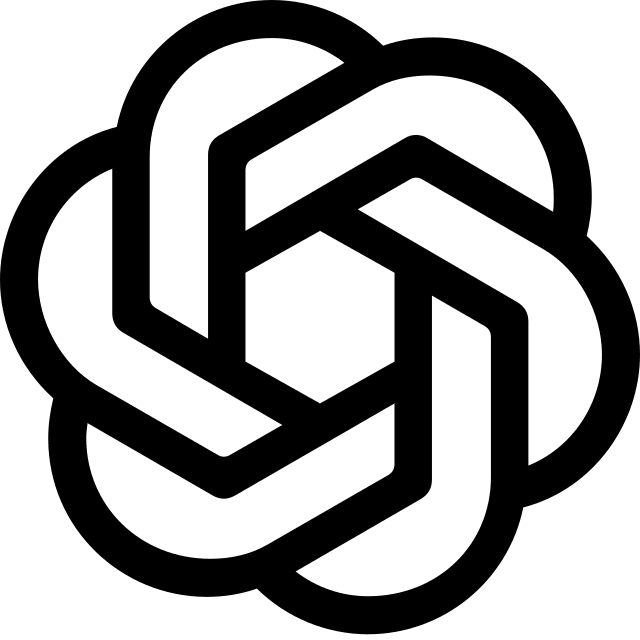}\hspace{0.35em}GPT-6 Astra & Codex & \textrm{high} & \textbf{77.34} & \textbf{80.8} & 95.7 & \textbf{24.02} & \textbf{30.02} & \textbf{23.30} & 321 \\
    \lbicon{openai}\hspace{0.35em}GPT-6 Astra & Codex & \textrm{xhigh} & 72.60 & 74.8 & \textbf{97.1} & 22.31 & 28.59 & 21.70 & 967 \\
    \lbicon{openai}\hspace{0.35em}GPT-6 Astra & Codex & \textrm{max} & 68.76 & 71.0 & 96.8 & 21.15 & 27.04 & 20.57 & 1,598 \\
    \lbicon{openai}\hspace{0.35em}GPT-6 Astra & Codex & \textrm{medium} & 68.16 & 71.6 & 95.2 & 21.25 & 26.36 & 20.54 & 208 \\
    \lbicon{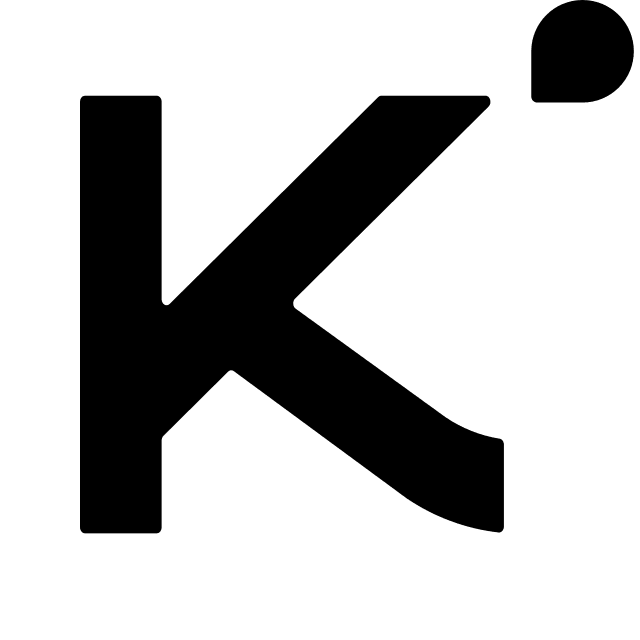}\hspace{0.35em}Kimi K3 & OpenCode & \textrm{high} & 67.80 & 73.8 & 91.9 & 21.49 & 26.33 & 19.97 & 1,158 \\
    \lbicon{openai}\hspace{0.35em}GPT-6 Astra & Codex & \textrm{low} & 59.42 & 63.0 & 94.3 & 18.65 & 22.99 & 17.78 & 179 \\
    \lbicon{openai}\hspace{0.35em}GPT-5.6 Sol & Codex & \textrm{max} & 49.28 & 52.6 & 93.7 & 15.57 & 18.93 & 14.78 & 988 \\
    \lbicon{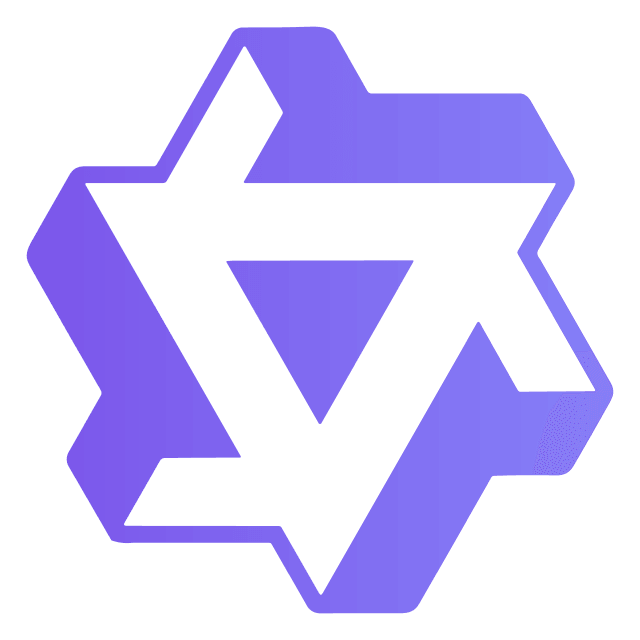}\hspace{0.35em}Qwen 3.8 Max & Claude Code & \textrm{xhigh} & 47.69 & 53.2 & 89.7 & 15.51 & 17.99 & 14.19 & 307 \\
    \lbicon{openai}\hspace{0.35em}GPT-5.6 Sol & Codex & \textrm{xhigh} & 36.24 & 40.0 & 90.6 & 11.73 & 13.50 & 11.01 & 495 \\
    \lbicon{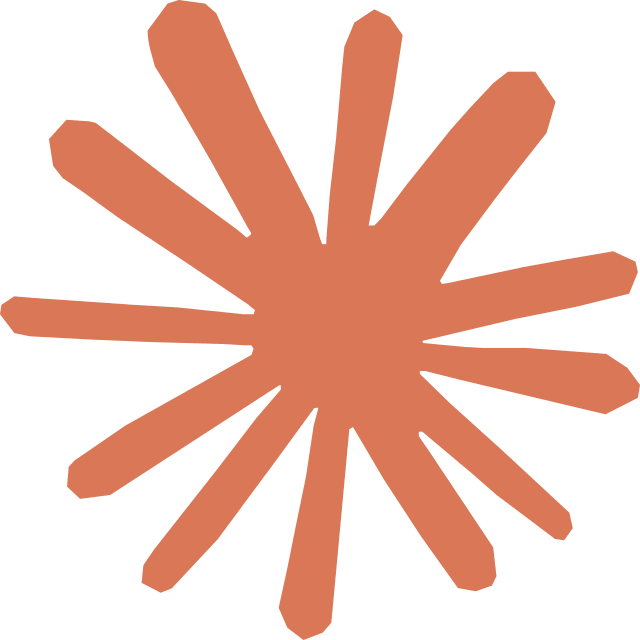}\hspace{0.35em}Claude Opus 5 & Claude Code & \textrm{xhigh} & 35.91 & 43.6 & 82.4 & 11.42 & 15.59 & 8.90 & 685 \\
    \lbicon{claude}\hspace{0.35em}Claude Opus 5 & Claude Code & \textrm{max} & 33.54 & 40.8 & 82.2 & 10.77 & 14.46 & 8.32 & 719 \\
    \lbicon{openai}\hspace{0.35em}GPT-5.6 Terra & Codex & \textrm{max} & 33.46 & 37.4 & 89.5 & 10.99 & 12.07 & 10.40 & 283 \\
    \lbicon{claude}\hspace{0.35em}Claude Opus 5 & Claude Code & \textrm{high} & 33.12 & 40.8 & 81.2 & 10.62 & 14.06 & 8.43 & 365 \\
    \lbicon{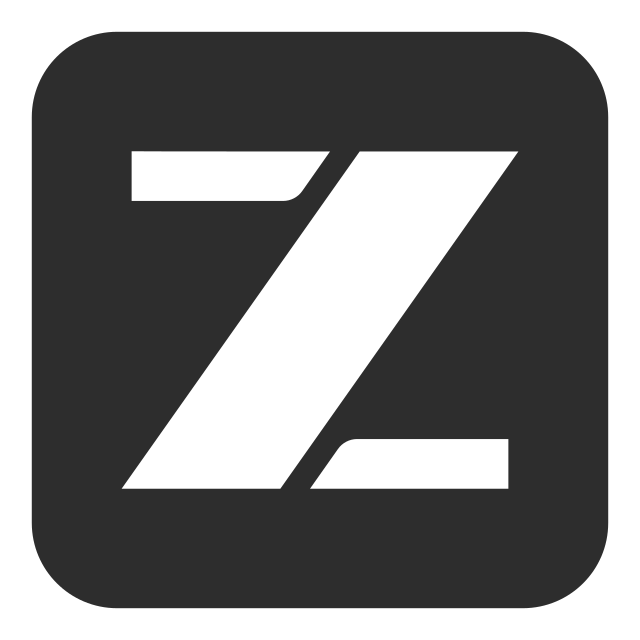}\hspace{0.35em}GLM-5.3-Flash & OpenCode & \textrm{max} & 32.42 & 36.8 & 88.1 & 10.45 & 12.61 & 9.36 & 53 \\
    \lbicon{openai}\hspace{0.35em}GPT-5.6 Luna & Codex & \textrm{max} & 31.02 & 36.4 & 85.2 & 10.70 & 10.43 & 9.89 & 166 \\
    \lbicon{openai}\hspace{0.35em}GPT-5.6 Sol & Codex & \textrm{high} & 28.28 & 31.8 & 88.9 & 9.28 & 10.27 & 8.73 & 285 \\
    \lbicon{openai}\hspace{0.35em}GPT-5.6 Terra & Codex & \textrm{xhigh} & 25.84 & 29.6 & 87.3 & 8.67 & 9.25 & 7.92 & 125 \\
    \lbicon{openai}\hspace{0.35em}GPT-5.6 Luna & Codex & \textrm{xhigh} & 24.63 & 29.0 & 84.9 & 8.52 & 8.27 & 7.83 & 87 \\
    \lbicon{claude}\hspace{0.35em}Claude Opus 5 & Claude Code & \textrm{medium} & 24.13 & 31.2 & 77.3 & 8.17 & 9.36 & 6.60 & 132 \\
    \lbicon{openai}\hspace{0.35em}GPT-5.6 Sol & Codex & \textrm{medium} & 22.20 & 25.4 & 87.4 & 7.45 & 8.08 & 6.67 & 179 \\
    \lbicon{openai}\hspace{0.35em}GPT-5.6 Luna & Codex & \textrm{high} & 19.47 & 23.8 & 81.8 & 6.89 & 6.44 & 6.15 & 57 \\
    \lbicon{claude}\hspace{0.35em}Claude Opus 5 & Claude Code & \textrm{low} & 17.90 & 23.6 & 75.8 & 6.02 & 6.89 & 4.98 & 73 \\
    \lbicon{openai}\hspace{0.35em}GPT-5.6 Terra & Codex & \textrm{high} & 16.89 & 19.4 & 87.0 & 5.67 & 6.08 & 5.14 & 81 \\
    \lbicon{openai}\hspace{0.35em}GPT-5.6 Sol & Codex & \textrm{low} & 11.12 & 13.0 & 85.6 & 3.80 & 3.88 & 3.45 & 123 \\
    \lbicon{openai}\hspace{0.35em}GPT-5.6 Terra & Codex & \textrm{medium} & 10.10 & 12.0 & 84.1 & 3.43 & 3.55 & 3.12 & 62 \\
    \lbicon{openai}\hspace{0.35em}GPT-5.5 & Codex & \textrm{medium} & 9.64 & 11.4 & 84.6 & 3.27 & 3.39 & 2.98 & 262 \\
    \lbicon{claude}\hspace{0.35em}Claude Sonnet 5 & Claude Code & \textrm{xhigh} & 8.98 & 12.8 & 70.1 & 2.75 & 3.97 & 2.26 & 214 \\
    \lbicon{claude}\hspace{0.35em}Claude Sonnet 5 & Claude Code & \textrm{high} & 8.76 & 12.4 & 70.7 & 2.82 & 3.74 & 2.20 & 137 \\
    \lbicon{openai}\hspace{0.35em}GPT-5.6 Terra & Codex & \textrm{low} & 8.60 & 9.8 & 87.8 & 2.87 & 3.14 & 2.60 & 58 \\
    \lbicon{openai}\hspace{0.35em}GPT-5.6 Luna & Codex & \textrm{medium} & 8.15 & 10.0 & 81.5 & 2.93 & 2.65 & 2.57 & 31 \\
    \lbicon{claude}\hspace{0.35em}Claude Sonnet 5 & Claude Code & \textrm{medium} & 5.99 & 9.2 & 65.1 & 1.80 & 2.65 & 1.54 & 99 \\
    \lbicon{openai}\hspace{0.35em}GPT-5.6 Sol & Codex & \textrm{none} & 4.99 & 5.8 & 86.0 & 1.65 & 1.88 & 1.46 & 117 \\
    \lbicon{claude}\hspace{0.35em}Claude Sonnet 5 & Claude Code & \textrm{low} & 4.03 & 6.4 & 63.0 & 1.16 & 1.79 & 1.08 & 61 \\
    \lbicon{openai}\hspace{0.35em}GPT-5.6 Terra & Codex & \textrm{none} & 3.55 & 4.2 & 84.4 & 1.20 & 1.32 & 1.03 & 45 \\
    \lbicon{openai}\hspace{0.35em}GPT-5.6 Luna & Codex & \textrm{low} & 3.03 & 3.6 & 84.3 & 1.02 & 1.08 & 0.93 & 21 \\
    \lbicon{openai}\hspace{0.35em}GPT-5.6 Luna & Codex & \textrm{none} & 1.53 & 1.8 & 85.3 & 0.50 & 0.55 & 0.48 & 22 \\
    \bottomrule
  \end{tabular}
  \end{adjustbox}
\end{table}

\textbf{Cost varies widely even among similarly performing systems.}
Reported generation cost for 500 tasks ranges from \$20.67 to \$1{,}598.36, with a median of
about \$151.50. GPT-6 Astra averages 156{,}884 tokens per task at high effort and
849{,}174 at max. Qwen~3.8~Max scores within 1.59 points of Sol Max for approximately one-third
of its cost (\$307 versus \$988). GLM-5.3-Flash reaches 32.42 for \$52.53, while GPT-6 Astra High costs about \$321 and Astra Max is the most expensive configuration. These are
comparisons among the evaluated model--harness configurations under the frozen price snapshot.

\subsection{Agent Behavior and Failure Modes}
\label{subsec:agent_analysis_main}

\begin{figure}[t]
  \centering
  \includegraphics[width=\textwidth]{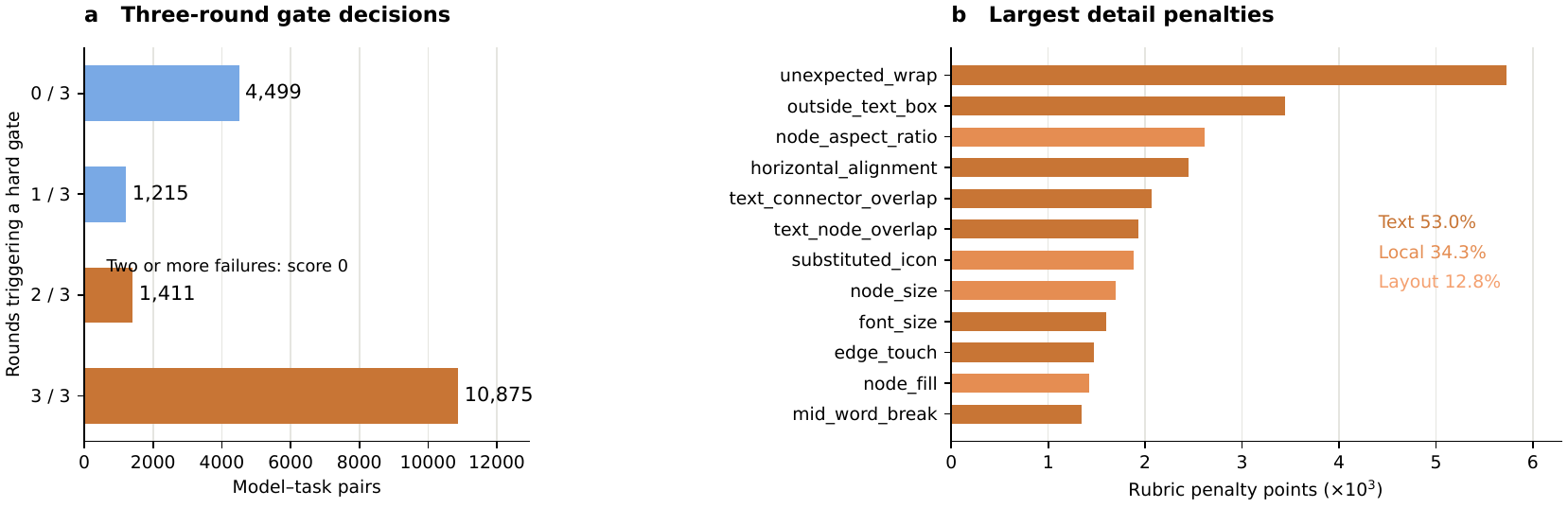}
  \caption{Failure analysis over 18{,}000 model--task pairs. Left: number of rounds triggering a hard gate. Right: the twelve largest detail penalties among gate-passing cases; dimension shares include all issue types.}
  \label{fig:main_failures}
\end{figure}

\textbf{Valid decks are common, but preserving the depicted process remains difficult.}
Across all 18{,}000 pairs, 5{,}714 (31.74\%) pass all gates. Figure~\ref{fig:main_failures}
shows how many rounds reject each reconstruction; two or more rejections produce a zero score.
The per-model breakdown in Appendix~\ref{app:outcomes} identifies semantics as the dominant
failure stage. A deck can preserve its palette and panel layout while reversing arrows or changing
branch targets (Appendix~\ref{subsec:analysis}). Gate-pass rates span 1.8--80.8\%, whereas mean
scores on passed tasks span 63.0--97.1. Mean score tracks gate passage closely ($r=0.997$),
consistent with the multiplicative score and the larger variation in gate success.

\textbf{Text fitting accounts for most residual deductions.}
Among gate-passing cases, text contributes 53.0\% of the 64{,}210.58 penalty points, local graphics
34.3\%, and layout 12.8\%. The largest individual deduction is \texttt{unexpected\_wrap}: 5{,}724
points across 1{,}864 cases, with a mean affected fraction of 0.577. Clipping, overflow, and
unintended line breaks expose a gap between declared text-frame geometry and rendered output.
The OOXML text frame specifies a container, while the renderer determines how its contents wrap.
Inspecting native objects alone can therefore miss defects that become visible after rendering.
Appendix~\ref{subsec:analysis} reports issue frequencies and representative gate failures.

\begin{figure}[t]
  \centering
  \includegraphics[width=\textwidth]{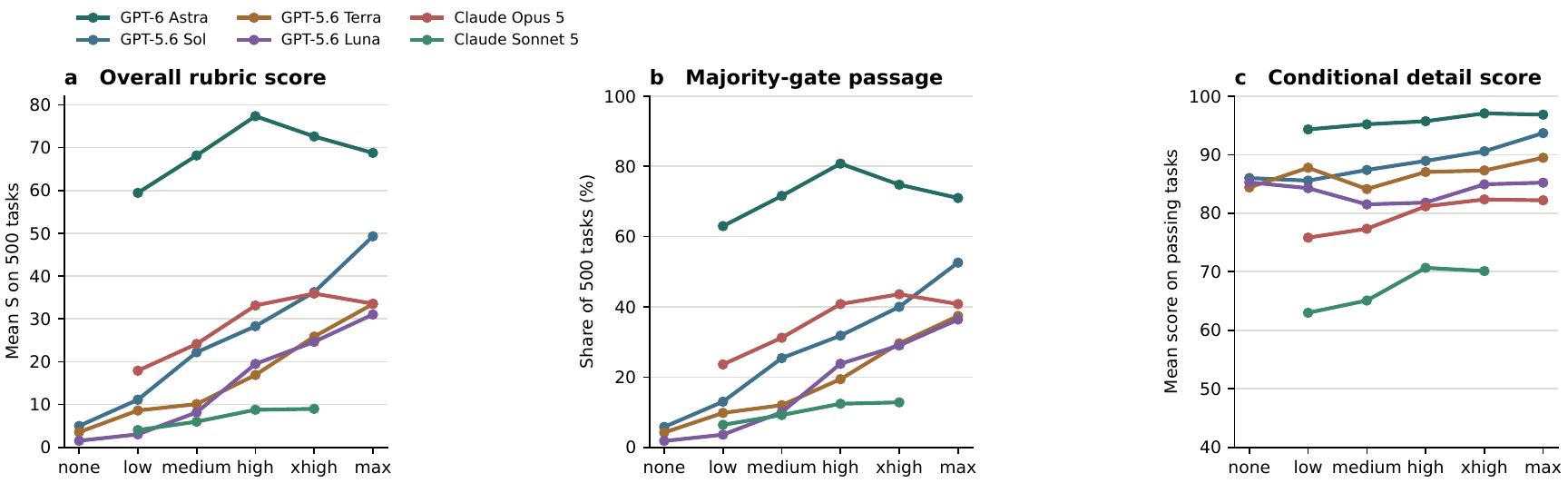}
  \caption{Reasoning-effort sweeps across six model families: mean score over all tasks (left), gate-pass rate (center), and mean score on passed tasks (right).}
  \label{fig:main_scaling}
\end{figure}

\textbf{More reasoning primarily improves gate passage, with diminishing or non-monotonic gains.}
GPT-6 Astra gains 17.92 points from low to high, with a 17.8-point gate-pass increase and only a
1.4-point rise in score on passed tasks. Sol gains 44.29 points from none to max as its gate-pass rate
rises by 46.8 points. Terra and Luna show the same broad pattern (Figure~\ref{fig:main_scaling}).
Additional effort is not uniformly beneficial: Opus peaks at xhigh (35.91) and falls to 33.54 at
max, while Sonnet changes from 8.76 at high to 8.98 at xhigh and loses artifact validity (98.4\%
to 95.6\%). More compute chiefly changes how often systems recover the process; it does not
consistently eliminate local visual defects.

Conditional detail means can involve different passing tasks. On the 289 tasks shared by
Astra High and Max, Max improves detail by 0.98 points while its overall gate-pass rate falls by
9.8 percentage points. Appendix~\ref{app:paired_analysis} reports these matched comparisons.

\textbf{Inspecting a reconstruction correlates with quality more strongly than editing it.}
For the 27 configurations with parsed trajectories, mean score correlates with review turns at
Pearson $r=0.881$ and Spearman $\rho=0.897$, versus $r=0.255$ and $\rho=0.462$ for revision
turns (Appendix~\ref{app:trajectories}, Figure~\ref{fig:process}). A review renders or inspects the candidate; a revision edits the
deck. This association is compatible with the text-fitting failures above, but does not establish
that forced inspection improves quality: reasoning effort affects both inspection and performance,
and equally frequent reviews yield different scores across families. For example, at max effort,
Sol averages 6.75 reviews per task and scores 49.28, while Luna averages 6.48 and scores 31.02;
Terra reviews 3.15 times and scores 33.46. Within-family review--score correlations range from
0.922 to 0.998, but revision counts also correlate positively within these effort sweeps
(Table~\ref{tab:trajectory_correlations}). The pooled contrast thus describes differences in
agent behavior without isolating the effect of visual inspection.

\subsection{Judge Agreement and Ranking Stability}
\label{subsec:validation_main}

\textbf{The judge substantially agrees with blinded human gate decisions.}
Three annotators independently label 200 cases, with four reconstructions per task across 50
tasks. Their decisions are unanimous in 81.5\% of cases. The three-round Luna High majority
agrees with the human majority on 86.5\% (Wilson 95\% CI 81.1--90.6\%), with 15 extra gates
and 12 missed gates (Table~\ref{tab:judge_validation}). Human unanimity and agreement between
majority votes measure different quantities, so the former is not an exact ceiling for the latter.
Appendix~\ref{app:validation_design} gives the sampling, blinding, and judge-selection procedure.

\begin{table}[t]
  \caption{Reported human agreement on 200 cases. Gate failure is the positive class.}
  \label{tab:judge_validation}
  \centering
  \footnotesize
  \setlength{\tabcolsep}{4pt}
  \renewcommand{\arraystretch}{1.02}
  \begin{minipage}[t]{0.42\linewidth}
    \vspace{0pt}
    \centering  
    \begin{tabular}{lcc}
      \toprule
      \multicolumn{3}{c}{\textbf{Human--human pairwise}} \\
      Pair & Gate agr. & Cohen's $\kappa$ \\
      \midrule
      Judge~1--Judge~2 & \textbf{90.5\%} & \textbf{0.781} \\
      Judge~1--Judge~3 & 89.0\% & 0.766 \\
      Judge~2--Judge~3 & 83.5\% & 0.645 \\
      \bottomrule
    \end{tabular}

    \vspace{0.4em}

    \begin{tabular}{lrr}
      \toprule
      \multicolumn{3}{c}{\textbf{Luna vs.\ human majority} ($N=200$)} \\
      & \multicolumn{2}{c}{Human majority} \\
      \cmidrule(l){2-3}
      Luna majority & Gate & Pass \\
      \midrule
      Gate & 120 & 15 \\
      Pass & 12 & 53 \\
      \bottomrule
    \end{tabular}
  \end{minipage}\hfill
  \begin{minipage}[t]{0.55\linewidth}
    \vspace{0pt}
    \centering
    \begin{tabular}{lr}
      \toprule
      \multicolumn{2}{c}{\textbf{Headline metrics}} \\
      Metric & Value \\
      \midrule
      Human three-way unanimity & 81.5\% \\
      Fleiss' $\kappa$ (three human raters) & 0.728 \\
      \addlinespace[2pt]
      Gate agreement & 86.5\% (173/200) \\
      Case-level Wilson 95\% CI & 81.1\%--90.6\% \\
      Cohen's $\kappa$ & 0.696 \\
      Precision (gate) & 88.9\% \\
      Recall (gate) & 90.9\% \\
      $F_1$ (gate) & 89.9\% \\
      \addlinespace[2pt]
      Luna extra gates (false positives) & 15 \\
      Luna missed gates (false negatives) & 12 \\
      \bottomrule
    \end{tabular}
  \end{minipage}
\end{table}

\textbf{Independent three-round majorities agree on 96.0\% of gate decisions.}
Luna High is unanimous across three individual rounds in 87.5\% of cases. Two independently
collected three-round majorities agree on 96.0\%, with 8 disagreements among 200 cases
(Table~\ref{tab:judge_robustness}). Appendix~\ref{app:detail_variation} separately measures
Stage~3 variation within Astra's recorded three-round evaluations.

\begin{table}[t]
  \caption{Repeatability of single-round and three-round gate decisions on 200 cases.}
  \label{tab:judge_robustness}
  \centering
  \footnotesize
  \setlength{\tabcolsep}{4pt}
  \renewcommand{\arraystretch}{1.02}
  \begin{minipage}[t]{0.53\linewidth}
    \vspace{0pt}
    \centering
    \begin{tabular}{lccc}
      \toprule
      \multicolumn{4}{c}{\textbf{Single-round consistency} (3 repeats)} \\
      Judge & Unanimous & Fleiss' $\kappa$ & Flipped \\
            & gate rate &                  & cases \\
      \midrule
      Luna High   & \textbf{87.5\%} & \textbf{0.809} & \textbf{25} \\
      Sol Medium  & 86.5\% & 0.797 & 27 \\
      Terra High  & 81.0\% & 0.714 & 38 \\
      \bottomrule
    \end{tabular}
  \end{minipage}\hfill
  \begin{minipage}[t]{0.45\linewidth}
    \vspace{0pt}
    \centering
    \begin{tabular}{lr}
      \toprule
      \multicolumn{2}{c}{\textbf{Three-round majority, repeated}} \\
      \multicolumn{2}{c}{Luna High, rounds 1--3 vs.\ 4--6} \\
      \midrule
      Agreement & 96.0\% \\
      Cohen's $\kappa$ & 0.908 \\
      Gate $F_1$ & 0.971 \\
      Disagreements & 8 / 200 \\
      \bottomrule
    \end{tabular}
  \end{minipage}
\end{table}

\textbf{Detail findings retain evidence from individual judge rounds.}
Single-round support is 6.8\% for \texttt{unexpected\_wrap} and 73.3\% for
\texttt{node\_spacing} (Appendix~\ref{app:issues}). Human issue-set agreement is reported
separately from gate agreement in Appendix~\ref{app:validation_design}.
Across four detail-scoring variants, rank correlations with the original ordering remain
$\tau=0.971$--$0.997$ (Appendix~\ref{app:scoring_sensitivity}).
Appendix~\ref{app:gate_policy} compares one-pass, majority, and unanimous voting on Astra.

\textbf{Five hundred tasks yield stable broad performance tiers.}
Across 10{,}000 paired task-bootstrap resamples, larger task sets stabilize the ordering
(Table~\ref{tab:bootstrap}). At 500 tasks, mean Kendall's $\tau$ is 0.956 and Astra High ranks
first in 97.4\% of resamples. Appendix~\ref{app:bootstrap} gives the resampling protocol
and paired confidence intervals for neighboring configurations.

\begin{table}[ht]
  \caption{Ranking stability from 10{,}000 paired task-bootstrap resamples per benchmark size.}
  \label{tab:bootstrap}
  \centering
  \footnotesize
  \setlength{\tabcolsep}{4pt}
  \renewcommand{\arraystretch}{1.02}
  \begin{adjustbox}{max width=\linewidth}
  \begin{tabular}{rccccccrr}
    \toprule
    & \multicolumn{2}{c}{Kendall's $\tau$}
      & \multicolumn{4}{c}{Exact top-$k$ set reproduced (\%)}
      & \multicolumn{2}{c}{95\% score CI width} \\
    \cmidrule(lr){2-3}\cmidrule(lr){4-7}
    Tasks & mean & 5--95\% & $k=1$ & $k=2$ & $k=3$ & $k=5$
      & mean & maximum \\
    \midrule
    50 & 0.871 & 0.829--0.911 & 60.0 & 39.1 & 26.6 & 66.4 & 18.7292 & 26.2842 \\
    100 & 0.907 & 0.873--0.937 & 74.8 & 53.9 & 33.3 & 83.6 & 13.2508 & 18.2886 \\
    200 & 0.932 & 0.905--0.956 & 87.7 & 69.0 & 37.3 & 95.4 & 9.3986 & 12.8205 \\
    300 & 0.944 & 0.921--0.965 & 93.1 & 77.4 & 41.7 & 98.7 & 7.6752 & 10.6089 \\
    400 & 0.951 & 0.927--0.971 & 96.1 & 83.4 & 44.6 & 99.4 & 6.6420 & 9.1836 \\
    500 & \textbf{0.956} & 0.933--0.975 & \textbf{97.4} & \textbf{88.0} & \textbf{46.2} & \textbf{99.8} & \textbf{5.9488} & \textbf{8.2862} \\
    \bottomrule
  \end{tabular}
  \end{adjustbox}
\end{table}

\FloatBarrier

\newpage
\section{Related Work}
\label{sec:related_work}

\paragraph{Presentation, poster and office documents.} Slide generation includes document-to-deck
systems~\citep{fu2022doc2ppt,sun2021d2s}, LLM agents~\citep{zheng2025pptagent,yang2026slidesgenbench,chen2026presentbench},
and scientific posters~\citep{pang2025paper2poster}. These tasks evaluate content, layout, and style
against open-ended targets. Office benchmarks specify edits to existing
files~\citep{guo2023pptc,wang2024officebench}. The earlier PPTBench~\citep{huang2025pptbench}
covers detection, understanding, modification, and generation, using task-specific correctness
metrics and a holistic generation score. Slide2Code~\citep{tang2025slidecoder}
directly evaluates reference-image reconstruction as editable PPTX and is the closest task setting.
Chart-to-code~\citep{yang2025chartmimic,wu2024plot2code,tang2025chart2code,zhang2026realchart2code,li2026geocodebench}
and screenshot-to-markup~\citep{si2024design2code} also fix the reference, with plotting code or
markup as output. Our PPTBench concentrates on scientific process diagrams, checks native-object
validity, and gates process and connector fidelity before applying localized visual deductions.
These stages distinguish failure to preserve the depicted process from errors in appearance.
Table~\ref{tab:comparison} in Appendix~\ref{app:related} compares these settings.

\paragraph{Understanding, parsing and generating diagrams.} Figure question answering tests
chart and document understanding~\citep{masry2022chartqa,mathew2022infographicvqa}, while
CharXiv~\citep{wang2024charxiv} evaluates charts from real scientific papers. Structured
prediction recovers diagram constituents and relations~\citep{kembhavi2016diagram,sun2022frdetr},
and flowchart benchmarks test reasoning over process structure~\citep{pan2024flowlearn,singh2024flowvqa}.
Diagram generation from text or papers~\citep{zala2024diagrammergpt,huang2026scifig} also motivates
structural evaluation through graph comparison~\citep{liang2025diagrameval} and inverse
parsing~\citep{zhang2026sciflowbench}. PPTBench supplies one reference image and evaluates both
its appearance and depicted process in the reconstructed slide. The process and connector gate
checks visible nodes, connections, routed segments, and arrowheads without requiring the OOXML
tree to match a reference graph.

Image-to-TikZ and SVG reconstruction~\citep{belouadi2024detikzify,rodriguez2025starvector} and
Image2Struct~\citep{roberts2024image2struct} also recover structured representations from visual
targets. PPTBench studies this reconstruction problem under native PowerPoint constraints, on
scientific diagrams whose visible process can fail even when the rendered layout is close.
Its contribution is the task collection and staged evaluation of that setting. The separation of
artifact, gate, and detail results makes this difference experimentally inspectable.

\paragraph{Model-based judges and agentic execution.} LLM and multimodal judges approximate
human evaluation~\citep{zheng2023judging,chen2024mllmjudge,lee2024prometheusvision}.
Agent-as-a-Judge adds tool use~\citep{zhuge2024agentjudge}. PPTBench's judge uses image-analysis
scripts and returns gates and localized findings for deterministic scoring, with reported gate agreement
against blinded human labels and independent repetitions (Section~\ref{subsec:validation_main}).
Evaluated agents combine reasoning and tool use~\citep{yao2023react}, code
execution~\citep{wang2024codeact}, and revision from feedback~\citep{shinn2023reflexion,madaan2023selfrefine}.
Appendix~\ref{app:related} provides extended comparisons.

\begin{samepage}
\section{Conclusion}
\label{sec:conclusion}

We introduced PPTBench to evaluate how coding agents translate visual targets into structured,
editable slides. Its 500 scientific flow-diagram tasks and four-stage judge assess artifact
validity, process semantics, rendering quality, and visual detail. Across ten models and 36
configurations, agents generally produce valid slides but struggle to preserve the depicted
process and visual structure, with the best mean score reaching 77.34. These findings highlight
the challenge of recovering objects and their relationships from visual input and expressing
them faithfully through code. Future work can investigate explicit structural reasoning and
verification through rendered feedback, and extend evaluation to richer visual artifacts such
as interfaces and 3D scenes. More broadly, PPTBench supports the development of agents that
turn visual intent into meaningful artifacts that people can inspect, edit, and reuse.

\par
\end{samepage}

\clearpage
\bibliography{ref}
\bibliographystyle{assets/plainnat}

\newpage

\appendix

\section{Related benchmark comparisons}
\label{app:related}
Table~\ref{tab:comparison} locates PPTBench among visual reconstruction, diagram parsing, and
presentation benchmarks. Native PPTX is an output constraint; topology gating is a scoring choice.

\begin{table}[H]
\caption{Related evaluation settings, output representations, and evaluation targets.}
\label{tab:comparison}
\centering
\small
\begin{tabular}{@{}p{0.29\linewidth}p{0.25\linewidth}p{0.40\linewidth}@{}}
\toprule
Setting & Representation & Evaluation target\\
\midrule
Chart reconstruction\newline\citep{yang2025chartmimic,wu2024plot2code,tang2025chart2code,zhang2026realchart2code}
& Plotting code & Reproduce chart appearance and numerical content.\\[4pt]
Visual code reconstruction\newline\citep{belouadi2024detikzify,rodriguez2025starvector,roberts2024image2struct}
& TikZ, SVG, or other source & Recover a rendered target in an executable representation.\\[4pt]
Diagram parsing and evaluation\newline\citep{kembhavi2016diagram,sun2022frdetr,liang2025diagrameval}
& Diagram constituents and relations & Recover or compare semantic structure.\\[4pt]
Slide generation and editing\newline\citep{zheng2025pptagent,yang2026slidesgenbench,chen2026presentbench,guo2023pptc}
& Slide decks or PPTX edits & Generate an open-ended design or execute specified edits.\\[4pt]
Reference-image slide reconstruction\newline\citep{tang2025slidecoder}
& Editable PPTX & Reconstruct a slide design; assess content, layout, execution, and visual fidelity.\\[4pt]
PPTBench~\citep{huang2025pptbench}
& Answers or PowerPoint API sequences & Evaluate detection, understanding, modification, and generation with task-specific metrics.\\[4pt]
PPTBench (ours) & Native PPTX slide & Reconstruct a fixed scientific diagram; check native-object validity, visible process fidelity, and visual defects.\\
\bottomrule
\end{tabular}
\end{table}

\paragraph{Visual reconstruction.} Chart-to-code
systems~\citep{yang2025chartmimic,wu2024plot2code,zhao2025chartcoder,tang2025chart2code,zhang2026realchart2code,yang2024matplotagent},
TikZ and SVG generation~\citep{belouadi2024automatikz,belouadi2024detikzify,rodriguez2025starvector,wu2025chat2svg,xing2025llm4svg,zou2024vgbench},
and screenshot-to-markup
methods~\citep{beltramelli2018pix2code,si2024design2code,laurencon2024websight,yun2024web2code,li2025sketch2code,xiao2025interaction2code}
recover executable representations. Image2Struct~\citep{roberts2024image2struct} evaluates the
rendered reconstruction. PPTBench additionally requires native slide objects and evaluates
process semantics.

\paragraph{Presentation and document authoring.}
Slide and poster generation~\citep{fu2022doc2ppt,sun2021d2s,zheng2025pptagent,yang2026slidesgenbench,chen2026presentbench,ge2025autopresent,cui2026deepslides,pan2026aeslides,pang2025paper2poster}
and layout synthesis~\citep{gupta2021layouttransformer,inoue2023layoutdm,lin2023layoutprompter,jia2024cole,yang2024posterllava}
evaluate open-ended designs. Office benchmarks specify operations on existing
files~\citep{guo2023pptc,zhang2024pptcr,li2023sheetcopilot,ma2024spreadsheetbench,wang2024officebench}.
Slide2Code~\citep{tang2025slidecoder} fixes a reference image and evaluates editable-slide
reconstruction. Its 900-example benchmark has three complexity tiers; the main experiment uses
300 examples. Evaluation compares content and position with the original PPTX, alongside execution,
CLIP, and SSIM. PPTBench shares the image-to-editable-slide setting but collects 500 scientific
flow diagrams without target PPTX files. Its staged records separate native-object validity,
reference-specific process and connector requirements, and visual penalties.

\paragraph{Distinction from the earlier PPTBench.}
The earlier PPTBench~\citep{huang2025pptbench} contains 4{,}439 instances in 11 subtasks drawn
from 958 government presentations. Tasks on existing slides combine screenshots with structured
JSON; modification and generation use 17 predefined PowerPoint APIs. Evaluation uses exact
matching for detection and post-edit JSON, answer accuracy for understanding, and a six-point
LLM rubric for generation. Its generation task includes screenshot inputs, overlapping with
visual reconstruction. Our PPTBench fixes 500 scientific diagram images
without target PPTX files or supplied object decompositions. Agents infer editable objects and
connections from the reference using their choice of authoring library. The evaluation checks
native-object validity and applies hard gates to process and connector fidelity and rendering
quality, then converts localized findings into deterministic visual deductions. This design
isolates failures in scientific process reconstruction from local visual defects.

\paragraph{Diagram understanding and generation.}
Figure and document QA~\citep{kahou2018figureqa,methani2020plotqa,masry2022chartqa,mathew2021docvqa,mathew2022infographicvqa,lee2023pix2struct,masry2023unichart,wang2024charxiv,roberts2024scifibench},
structured parsing~\citep{kembhavi2016diagram,kembhavi2017tqa,sun2022frdetr,zhang2022pgdp,lu2021intergps},
and flowchart reasoning~\citep{tannert2023flowchartqa,pan2024flowlearn,singh2024flowvqa}
measure complementary forms of visual understanding. Diagram
generation~\citep{zala2024diagrammergpt,huang2026scifig,zhao2026crafter,guo2025paper2sysarch}
and structural evaluation~\citep{liang2025diagrameval,zhang2026sciflowbench}
address semantic fidelity. PPTBench evaluates the process depicted by editable objects without
requiring a unique reference OOXML tree.

\paragraph{Judges and coding agents.}
Model-based evaluation studies rubric adherence and judge
biases~\citep{zheng2023judging,chi2026frontier,liu2023geval,kim2024prometheus,chi2026ai4ai,zhu2023judgelm,panickssery2024selfpreference,hong2026swe,chen2024mllmjudge,lee2024prometheusvision,chen2024mjbench,li2025vlrewardbench}.
Agent-as-a-Judge introduces evaluator tool use~\citep{zhuge2024agentjudge}. PPTBench combines
this capability with deterministic scoring and human agreement checks~\citep{cohen1960kappa}.
Its tested agents use tool actions, code execution, and feedback-driven
revision~\citep{yao2023react,wang2024codeact,shinn2023reflexion,madaan2023selfrefine}; execution-based
benchmarks~\citep{jimenez2024swebench,xie2024osworld} and reasoning
scaling~\citep{wei2022cot,snell2024scaling} provide related evaluation settings.

\clearpage
\section{Task catalog}
\label{app:task_catalog}
Table~\ref{tab:representatives} lists domain counts and representative source papers.
The ten presentation domains aggregate 50 arXiv primary categories; source identifiers are
retained for traceability and withheld from agents.
\begin{table}[H]
  \caption{Task counts, arXiv categories, and representative source papers for the ten benchmark domains.}
  \label{tab:representatives}
  \centering
  \scriptsize
  \renewcommand{\arraystretch}{1.08}
  \begin{adjustbox}{max width=\linewidth}
  \begin{tabular}{p{0.20\linewidth}rp{0.32\linewidth}p{0.38\linewidth}}
    \toprule
    Domain & Tasks & arXiv classes & Representative source paper \\
    \midrule
    Systems, architecture and SE & 91 & \texttt{cs.SE (32); cs.DC (31); cs.AR (25); cs.CE (1); cs.ET (1); cs.OS (1)} & TDFlow: Agentic Workflows for Test Driven Development (arXiv: \texttt{2510.23761}) \\
    AI and machine learning & 90 & \texttt{cs.LG (54); cs.AI (19); cs.NE (15); stat.ML (2)} & A Distributionally Robust Framework for Nuisance in Causal Effect Estimation (arXiv: \texttt{2505.17717}) \\
    Computer vision and imaging & 67 & \texttt{cs.CV (51); eess.IV (15); cs.MM (1)} & Building Optimal Neural Architectures using Interpretable Knowledge (arXiv: \texttt{2403.13293}) \\
    NLP, speech and audio & 54 & \texttt{cs.CL (29); eess.AS (17); cs.SD (8)} & Speed Always Wins: A Survey on Efficient Architectures for Large Language Models (arXiv: \texttt{2508.09834}) \\
    Quantum and fundamental physics & 43 & \texttt{quant-ph (34); gr-qc (7); hep-ex (1); nucl-th (1)} & Pinball: A Cryogenic Predecoder for Surface Code Decoding Under Circuit-Level Noise (arXiv: \texttt{2512.09807}) \\
    Astronomy and astrophysics & 37 & \texttt{astro-ph.IM (30); astro-ph.EP (3); astro-ph.GA (2); astro-ph.HE (2)} & Validation of the HERA Phase I Epoch of Reionization 21 cm Power Spectrum Software Pipeline (arXiv: \texttt{2104.09547}) \\
    Robotics and HCI & 29 & \texttt{cs.RO (18); cs.HC (11)} & Implementing a Robot Intrusion Prevention System (RIPS) for ROS 2 (arXiv: \texttt{2412.19272}) \\
    Networking, security and IR & 25 & \texttt{cs.NI (10); cs.CR (5); cs.IT (5); cs.IR (4); cs.SI (1)} & An OPC UA-based industrial Big Data architecture (arXiv: \texttt{2306.01418}) \\
    Biomedicine and life sciences & 19 & \texttt{q-bio.QM (6); q-bio.NC (5); q-bio.BM (4); physics.med-ph (2); physics.bio-ph (1); q-bio.GN (1)} & MegaFold: Efficient Training of Next-Generation 3D Attention Protein Models on Cross-Platform GPUs (arXiv: \texttt{2506.20686}) \\
    Others & 45 & \texttt{physics.ins-det (2); math.DS (1); physics.comp-ph (1); physics.flu-dyn (1); stat.AP (1); cs.CY (5); econ.GN (3); physics.soc-ph (2); cond-mat.mtrl-sci (10); cond-mat.dis-nn (1); cond-mat.str-el (1); eess.SP (16); math.OC (1)} & A Flexible Data Acquisition System Architecture for the Nab Experiment (arXiv: \texttt{2407.17606}) \\
    \bottomrule
  \end{tabular}
  \end{adjustbox}
\end{table}

\clearpage
\section{Complete results and execution details}
\label{app:full_experiments}
\subsection{Execution and cost accounting}
\label{subsec:setup}

The generation prompt asks the agent to inspect the reference image and reconstruct it as a
single editable PowerPoint slide. It lists the approved raster resources for the task, requires
each resource to appear unmodified in its intended location, and requires all other diagram
content to use native PowerPoint objects. Agents choose their authoring method; python-pptx and
Pillow are provided. Before submission, the agent must open or parse the deck to verify its validity.

Astra uses Codex CLI 0.154.0, with generation concurrency 10 and judge concurrency 4 per
effort. Model-call, first-event, and stream-idle timeouts are disabled. Invalid or missing model
outputs score zero. Rendering operates on the final generated deck. These runtime settings
describe Astra; the other model--harness combinations are listed in Table~\ref{tab:main_leaderboard}.

We report the final generation-cost total for each configuration over the 500-task evaluation,
excluding judging. Costs use recorded provider charges or token usage repriced at a frozen
price snapshot, with reasoning tokens billed as output and cached input accounted for separately.
For Astra, the estimate uses the generation associated with each final evaluated output,
including artifact failures. Rates per million tokens are \$10 for uncached input, \$1 for
cached input, \$12.50 for cache writes, and \$50 for output; long-context surcharges,
service-tier adjustments, and tool fees are excluded. The other configurations use their
recorded generation totals. Astra averages 156{,}884 tokens per task at high effort and
849{,}174 at max.

\subsection{Image preparation and evaluation}
\label{subsec:package}

Evaluation applies Stage~0 artifact validation, renders the candidate with LibreOffice,
runs the three judge stages, and aggregates three rounds with deterministic scoring rules.

Normalization uses the median border color as background, a content threshold of 10 intensity
levels, and a 2\% crop margin with a four-pixel minimum. Aspect-preserving crops are centered in
$1200\times900$ white panels separated by four pixels. Full-resolution images accompany the pair.
All 2{,}500 Astra runs exclude captions. Agents receive reference and resource images without
source-paper provenance.

\subsection{Outcome proportions}
\label{app:outcomes}

\begin{figure}[H]
  \centering
  \includegraphics[width=\textwidth]{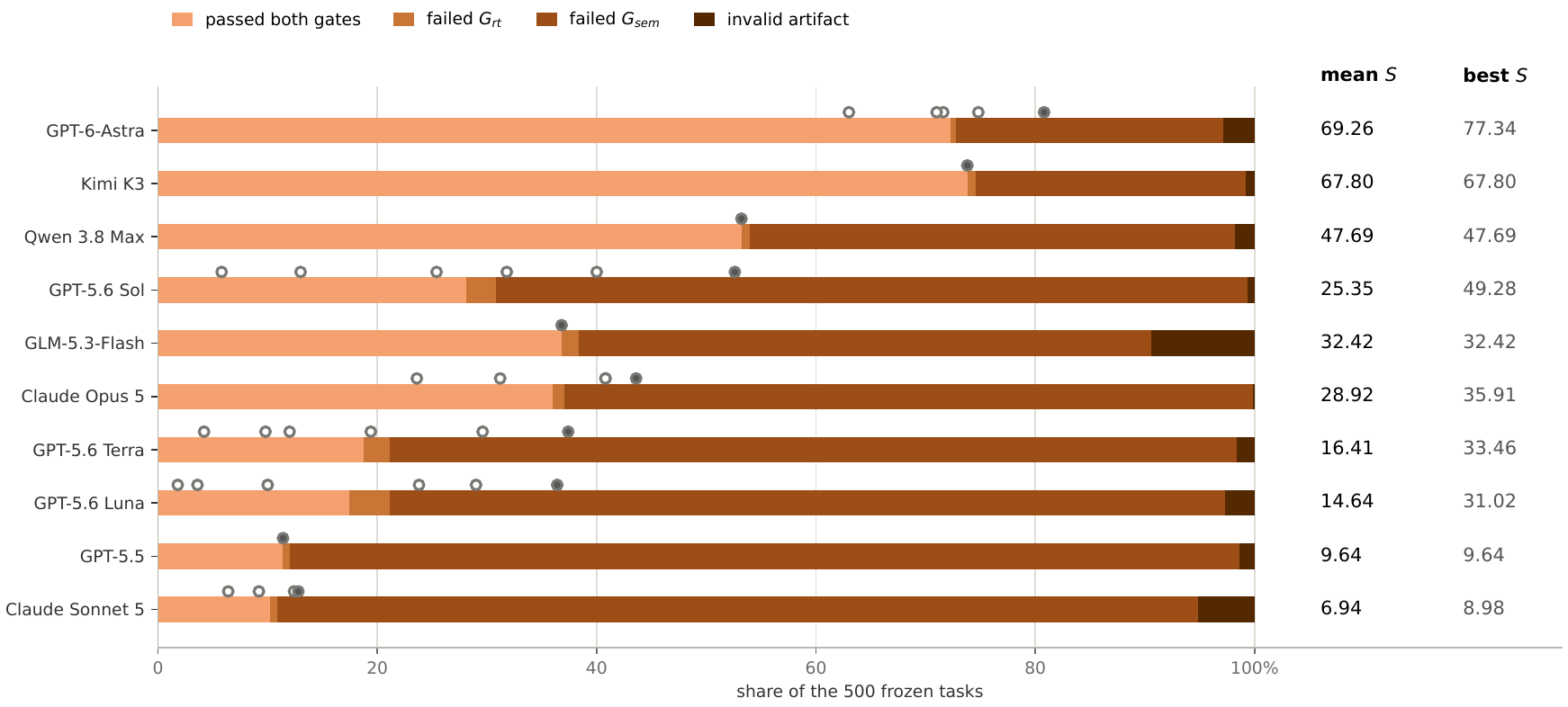}
  \caption{Outcome proportions by model, averaged across tested effort settings. Dots mark the
  gate-pass rates of individual configurations.}
  \label{fig:results}
\end{figure}

Figure~\ref{fig:results} averages outcome proportions equally across each model's tested effort
settings. Across all 36 configurations, 5{,}714 cases pass all gates, including 461 with no
recorded detail finding. These model-level averages complement Table~\ref{tab:main_leaderboard}.

\clearpage
\section{Failure cases and detail statistics}
\label{subsec:analysis}

\begin{figure}[H]
  \centering
  \includegraphics[width=\textwidth]{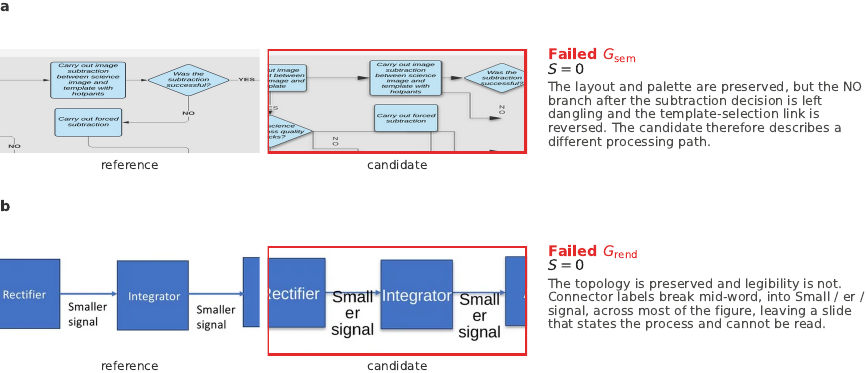}
  \caption{Examples of hard-gate failures: (a) incorrect process connections from Astra High
  on task~0017; (b) illegible connector labels from Luna Max. Both fail all three judge rounds.}
  \label{fig:gate_failures}
\end{figure}

\begin{figure}[H]
  \centering
  \includegraphics[width=\textwidth]{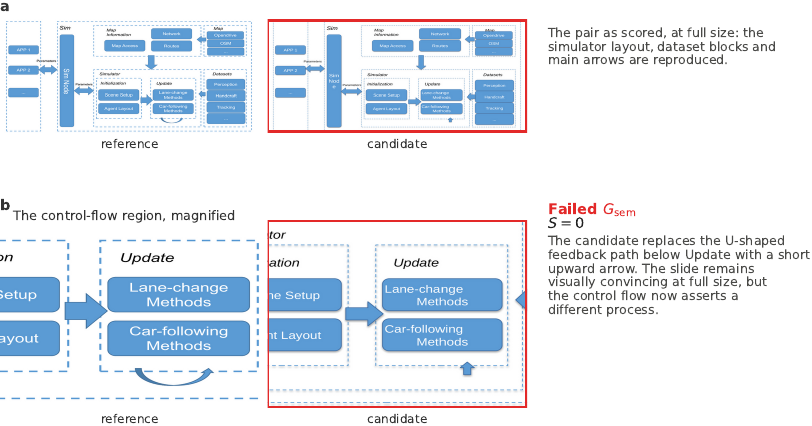}
  \caption{Semantic failure on task~0104 (Astra High): (a) full reference and reconstruction;
  (b) enlarged control-flow region showing the altered feedback connection.}
  \label{fig:gate_semantic}
\end{figure}

Figures~\ref{fig:gate_failures} and~\ref{fig:gate_semantic} show process/connector and rendering
failures. Task~0017 has a disconnected subtraction branch and rerouted selection path; task~0104
has an altered feedback connector below the Update block.


\clearpage
\subsection{Issue frequencies and round support}
\label{app:issues}

Table~\ref{tab:issues} reports the fifteen largest penalties among the 5{,}714 gate-passing
reconstructions. The Cases column counts model--task pairs with at least one occurrence;
Category ratio averages the largest affected fraction across passing rounds for each case.
Single-round support is the fraction of cases in which only one passing round records
the issue. Across all issue types, text and local graphics account for 10{,}191 and 10{,}849
recorded occurrences, respectively.

\begin{table}[H]
  \caption{The fifteen largest aggregate detail penalties among 5{,}714 gate-passing reconstructions. Cases may contain multiple issue types.}
  \label{tab:issues}
  \centering
  \small
  \begin{tabular}{lllrrrr}
    \toprule
    Issue type & Dimension & Severity & Cases & Category
      & Total & Single- \\
    & & & & ratio & penalty & round (\%) \\
    \midrule
    \texttt{unexpected\_wrap} & Text & Medium & 1,864 & 0.577 & 5,723.9 & 6.8 \\
    \texttt{outside\_text\_box} & Text & High & 658 & 0.319 & 3,441.8 & 35.7 \\
    \texttt{node\_aspect\_ratio} & Graphics & Medium & 802 & 0.655 & 2,614.5 & 53.7 \\
    \texttt{horizontal\_alignment} & Text & Medium & 737 & 0.694 & 2,446.7 & 42.1 \\
    \texttt{text\_connector\_overlap} & Text & High & 385 & 0.342 & 2,061.8 & 55.3 \\
    \texttt{text\_node\_overlap} & Text & High & 379 & 0.306 & 1,929.8 & 43.5 \\
    \texttt{substituted\_icon} & Graphics & Medium & 542 & 0.794 & 1,877.0 & 36.5 \\
    \texttt{node\_size} & Graphics & Medium & 520 & 0.654 & 1,699.7 & 67.1 \\
    \texttt{font\_size} & Text & Low & 1,315 & 0.905 & 1,593.2 & 46.1 \\
    \texttt{edge\_touch} & Text & Medium & 574 & 0.308 & 1,473.1 & 62.7 \\
    \texttt{node\_fill} & Graphics & Low & 1,200 & 0.848 & 1,421.2 & 44.2 \\
    \texttt{mid\_word\_break} & Text & Medium & 434 & 0.575 & 1,343.5 & 48.2 \\
    \texttt{vertical\_alignment} & Text & Medium & 382 & 0.798 & 1,325.8 & 76.2 \\
    \texttt{text\_color} & Text & Medium & 388 & 0.737 & 1,307.0 & 45.1 \\
    \texttt{node\_spacing} & Layout & Medium & 360 & 0.781 & 1,237.6 & 73.3 \\
    \bottomrule
  \end{tabular}
\end{table}

\clearpage
\section{Matched reasoning and configuration weighting}
\label{app:paired_analysis}

\paragraph{Common-pass comparisons.}
For each effort pair, we retain tasks that pass under both settings and subtract the lower-effort
detail score from the higher-effort score. Table~\ref{tab:paired_reasoning} compares high with
the highest available effort. Pass-rate differences use all 500 tasks. Intervals are percentile
intervals from 10{,}000 paired task resamples, seed 20260925, with no multiplicity adjustment.
Generated outputs and judge findings remain fixed.

\begin{table}[H]
\centering
\small
\caption{High versus highest effort on shared passing tasks. $N$ is common-pass count; $\Delta p$ uses all 500 tasks.}
\label{tab:paired_reasoning}
\setlength{\tabcolsep}{5pt}
\begin{tabular}{llrrrr}
\toprule
Model & Higher effort & $N$ & $\Delta$ detail & 95\% CI & $\Delta p$ (pp)\\
\midrule
Opus 5 & max & 147 & 1.76 & [0.15, 3.43] & 0.0\\
Sonnet 5 & xhigh & 35 & 0.58 & [-2.75, 3.87] & 0.4\\
Luna & max & 85 & 4.39 & [1.88, 7.01] & 12.6\\
Sol & max & 109 & 4.13 & [2.43, 5.91] & 20.8\\
Terra & max & 66 & 2.83 & [0.22, 5.47] & 18.0\\
Astra & max & 289 & 0.98 & [0.18, 1.75] & -9.8\\
\bottomrule
\end{tabular}
\end{table}

For Astra High versus Max, 289 tasks pass at both settings. Max gains 66 passing tasks and loses
115, giving a net decrease of 49 out of 500 tasks, or 9.8 percentage points. On the shared
passing tasks, detail increases by 0.98 points. This distinguishes changes in the passing set
from changes in reconstruction quality (Section~\ref{subsec:agent_analysis_main}).

\paragraph{Configuration weighting.}
Equal-family weighting first averages available effort settings within each family, then averages
the ten family scores. Highest-effort selection retains one configuration per family; available
effort levels differ. Observed-best selection chooses each family's highest-scoring setting on
these same 500 tasks.
Table~\ref{tab:menu_sensitivity} quantifies these choices. Its confidence intervals use paired
500-task resamples.

\begin{table}[H]
\centering
\small
\caption{Aggregate score under alternative configuration menus. Median is over configuration or family means, as specified.}
\label{tab:menu_sensitivity}
\begin{tabular}{lrrr}
\toprule
Weighting / selection & Mean & 95\% CI & Median\\
\midrule
36 configurations, equal weight & 28.18 & [26.72, 29.62] & 24.38\\
10 families, mean over available efforts & 31.91 & [30.40, 33.48] & 27.14\\
10 families, highest available effort & 38.26 & [36.39, 40.10] & 33.50\\
10 families, best observed effort & 39.36 & [37.49, 41.25] & 34.69\\
\bottomrule
\end{tabular}
\end{table}

\clearpage
\section{Generation trajectories}
\label{app:trajectories}

A review turn renders or inspects the candidate; a revision turn edits the deck. The trajectory
analysis covers 27 configurations in total: 26 across five effort sweeps and one GPT-5.5 configuration
(Table~\ref{tab:trajectory_correlations}). The covered families are Sol, Terra, Luna, Opus,
Sonnet, and GPT-5.5; Sol Medium is excluded. Correlations summarize the observed effort sweeps,
in which reasoning effort varies alongside inspection, revision, and score.

\begin{figure}[H]
  \centering
  \includegraphics[width=\textwidth]{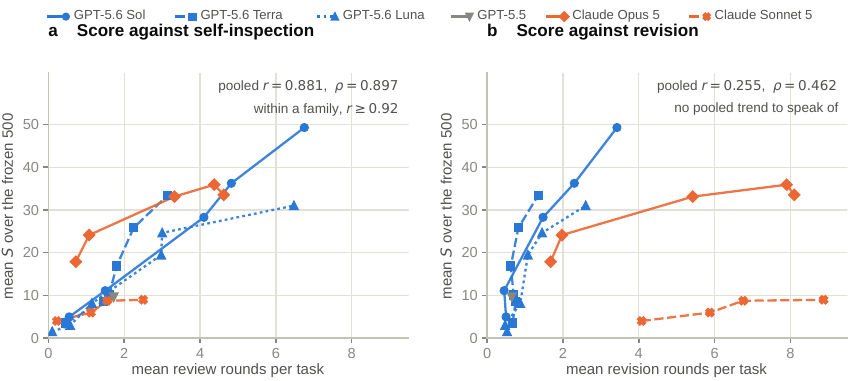}
  \caption{Mean score versus review turns (left) and revision turns (right) for 27 configurations.
  Connected points follow each model's reasoning-effort sweep.}
  \label{fig:process}
\end{figure}
\begin{table}[H]
  \caption{Correlations between mean score and generation behavior. A dash indicates an
  undefined within-model correlation for the single GPT-5.5 configuration.}
  \label{tab:trajectory_correlations}
  \centering
  \small
  \begin{tabular}{lrrrrr}
    \toprule
    & & \multicolumn{2}{c}{Review turns} & \multicolumn{2}{c}{Revision turns} \\
    \cmidrule(lr){3-4}\cmidrule(lr){5-6}
    Model & Configurations & Pearson $r$ & Spearman $\rho$ & Pearson $r$ & Spearman $\rho$ \\
    \midrule
    All matched configurations & 27 & 0.881 & 0.897 & 0.255 & 0.462 \\
    Claude Opus 5 & 5 & 0.947 & 0.900 & 0.927 & 0.900 \\
    Claude Sonnet 5 & 4 & 0.922 & 1.000 & 0.913 & 1.000 \\
    GPT-5.6 Luna & 6 & 0.946 & 1.000 & 0.920 & 0.943 \\
    GPT-5.6 Sol & 5 & 0.998 & 1.000 & 0.979 & 0.900 \\
    GPT-5.6 Terra & 6 & 0.955 & 1.000 & 0.798 & 0.600 \\
    GPT-5.5 & 1 & -- & -- & -- & -- \\
    \bottomrule
  \end{tabular}
\end{table}

\clearpage
\section{Judge reliability and ranking uncertainty}
\label{subsec:validity}

\subsection{Validation design and judge repeatability}
\label{app:validation_design}

\begin{figure}[H]
  \centering
  \includegraphics[width=\textwidth]{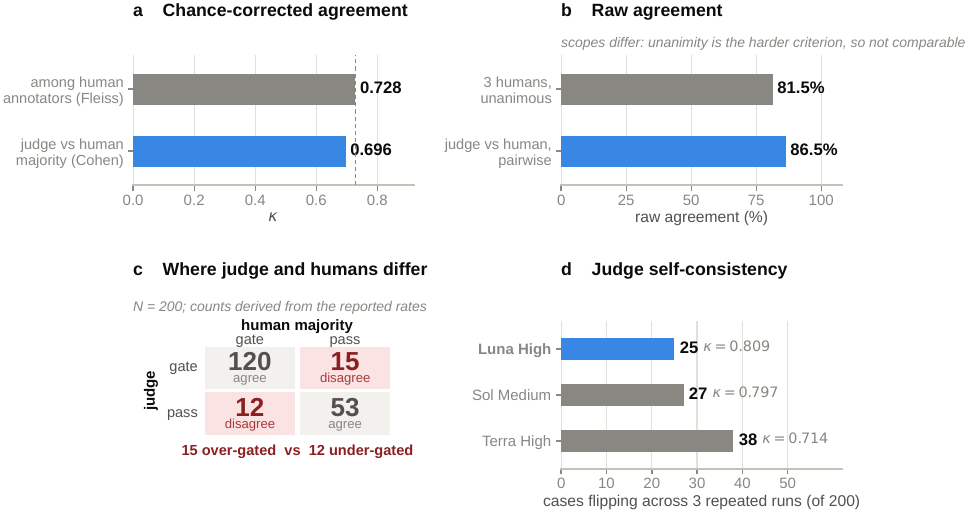}
  \caption{Gate validation on 200 cases: (a) chance-corrected agreement; (b) raw agreement;
  (c) confusion matrix; (d) single-round gate consistency across three repeats.}
  \label{fig:validity}
\end{figure}

Protocol development uses 50 tasks. Judge selection uses a disjoint set of 50 tasks, with four
reconstructions spanning quality per task, giving 200 validation cases. Three annotators
independently label these cases, blinded to model identity, reasoning effort, judge outputs,
and one another's labels. Both subsets belong to the 500-task evaluation set.
Table~\ref{tab:judge_validation} reports agreement on the judge-selection cases, and
Figure~\ref{fig:validity} summarizes gate agreement and repeatability. The Wilson interval is
computed over the 200 cases from the reported aggregate counts.

Prompts, taxonomy, severity bases, dimension caps, ratio staircase, and aggregation were frozen
before the initial leaderboard recomputation and retained for subsequent configurations.

Human gate agreement and issue-set agreement measure different decisions. On the 52 cases
passed by all three annotators, their issue sets match exactly in 9 cases (17.3\%), with mean
pairwise Jaccard similarity 0.392. Human validation therefore covers gates and issue sets;
continuous detail scores are examined through the judge-variation and scoring-policy analyses below.

Table~\ref{tab:judge_robustness} compares independent runs and two Luna High triplets.
Majorities from rounds~1--3 and independently collected rounds~4--6 agree on 192/200 cases;
three are gated only by the first triplet and five only by the second. The additional triplet
evaluates gates without Stage~3 enumeration.

\clearpage
\subsection{Within-triplet detail variation}
\label{app:detail_variation}

We analyze the three recorded judge rounds for each of Astra's 2{,}500 artifacts.
Reapplying consensus aggregation reproduces all reported scores exactly. Table~\ref{tab:detail_variation}
uses the 1{,}645 artifacts whose three rounds all reach Stage~3. Each round is scored separately
using its own findings and category ratios. Range is the maximum minus minimum of these three
scores; pairwise difference averages the three absolute score differences per artifact.

\begin{table}[H]
\centering
\small
\caption{Detail-score variation within existing Astra judge triplets, conditional on all three rounds passing.}
\label{tab:detail_variation}
\begin{tabular}{lrrrr}
\toprule
Effort & Artifacts & Median range & 95th percentile & Mean pairwise difference\\
\midrule
Low & 279 & 2.50 & 8.52 & 2.03\\
Medium & 324 & 1.87 & 7.76 & 1.84\\
High & 366 & 1.50 & 7.50 & 1.62\\
XHigh & 354 & 1.06 & 7.50 & 1.30\\
Max & 322 & 0.62 & 7.50 & 1.27\\
\midrule
All five & 1{,}645 & 1.25 & 7.50 & 1.60\\
\bottomrule
\end{tabular}
\end{table}

These measurements describe variation within a recorded triplet, conditional on all three
rounds passing. The independently repeated triplets in Table~\ref{tab:judge_robustness}
instead measure the repeatability of aggregate gate decisions.

\subsection{Sensitivity to gate-vote threshold}
\label{app:gate_policy}

We reaggregate Astra's 7{,}500 recorded rounds, requiring one, two, or three rounds to pass both
judged gates. Each policy retains the original union/max detail aggregation over non-gated rounds;
all 71 deterministic artifact failures remain zero. Majority reaggregation exactly reproduces
all 2{,}500 recorded results. Gate votes disagree in 258 cases across 170 task IDs. Requiring one
pass admits 97 majority failures; requiring unanimity rejects 161 majority passes.

Table~\ref{tab:gate_policy} reports the resulting scores and passing counts.
\begin{table}[htbp]
\centering
\small
\caption{Astra scores under three gate-vote thresholds. Parentheses give passing cases out of 500.}
\label{tab:gate_policy}
\begin{tabular}{lrrr}
\toprule
Effort & At least one pass & Majority (reported) & Unanimous pass\\
\midrule
Low & 63.15 (335) & 59.42 (315) & 52.81 (279)\\
Medium & 72.60 (382) & 68.16 (358) & 61.78 (324)\\
High & 80.56 (421) & 77.34 (404) & 70.20 (366)\\
XHigh & 75.94 (392) & 72.60 (374) & 68.85 (354)\\
Max & 72.10 (373) & 68.76 (355) & 62.59 (322)\\
\bottomrule
\end{tabular}
\end{table}

The per-case unanimity score is the recorded majority score multiplied by an indicator for
three passing rounds, so requiring unanimity can only preserve or decrease each score.
One-pass voting can admit cases rejected by the majority.

\subsection{Sensitivity to detail scoring policy}
\label{app:scoring_sensitivity}

We recompute 18{,}000 cases from the frozen merged findings, reproducing all original scores
before changing one policy at a time. Hard-gate outcomes remain fixed. Two-round support discards
issues observed in fewer than two non-gated rounds. Unit multiplier sets $m(\rho)=1$ while
retaining severity bases. Equal weights averages the three retained dimension fractions and
rescales to 100. Category maximum retains the largest subtype penalty within each dimension and
category, then applies the original cap. This variant measures sensitivity to accumulating
multiple subtype deductions within one category.

Table~\ref{tab:scoring_sensitivity} compares each variant with the original configuration ordering.

\begin{table}[H]
\centering
\small
\caption{Sensitivity of 36 configuration scores to detail policies. $\Delta$ measures score change from the original rubric.}
\label{tab:scoring_sensitivity}
\begin{tabular}{lrrrr}
\toprule
Policy & Kendall $\tau$ & Median $\Delta$ & Max. $|\Delta|$ & Reversed pairs\\
\midrule
Two-round issue support & 0.975 & 1.77 & 4.53 & 8/630\\
Unit affected-ratio multiplier & 0.971 & 1.70 & 4.15 & 9/630\\
Equal dimension weights & 0.997 & 0.13 & 0.55 & 1/630\\
Maximum per issue category & 0.981 & 1.21 & 2.74 & 6/630\\
\bottomrule
\end{tabular}
\end{table}

Among the 5{,}714 passed cases, two-round support raises mean detail by 5.89 points and
category maximum by 3.87. All four variants retain Astra High in first place and preserve
the top-five set. These comparisons hold the generated artifacts and judge findings fixed.

\clearpage
\subsection{Paired task bootstrap}
\label{app:bootstrap}

\begin{figure}[H]
  \centering
  \includegraphics[width=\textwidth]{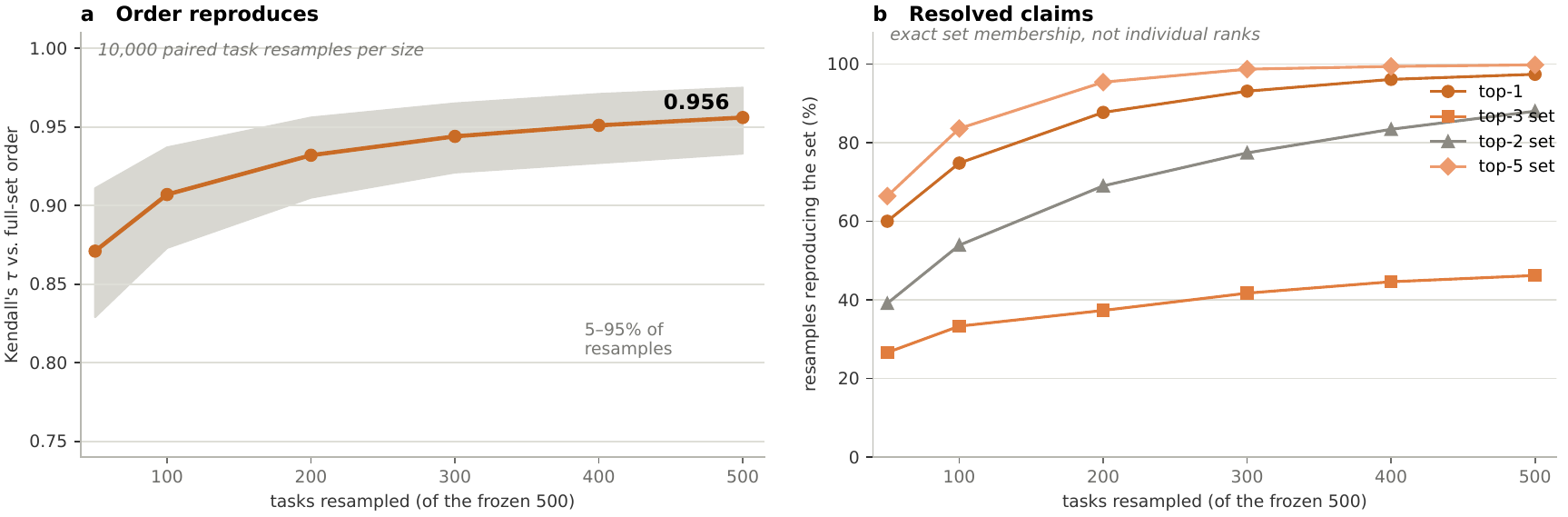}
  \caption{Ranking stability under paired task resampling: (a) Kendall's $\tau$, with a
  5--95\% band; (b) exact top-$k$ set recovery rates.}
  \label{fig:stability}
\end{figure}

At each benchmark size, we draw 10{,}000 task samples with replacement and evaluate every
configuration on the same sample. Table~\ref{tab:bootstrap} gives rank correlation, exact
top-$k$ membership, and score-interval widths; Figure~\ref{fig:stability} plots their dependence
on sample size. At 500 tasks, 5 of 35 adjacent score differences have 95\% intervals excluding
zero. Astra High's lead over Astra XHigh is 4.74 points, with paired 95\% CI $[-0.07,9.53]$.

\clearpage
\section{Scored reconstruction examples}
\label{app:case_study}
\begin{figure}[H]
  \centering
  \includegraphics[width=\textwidth]{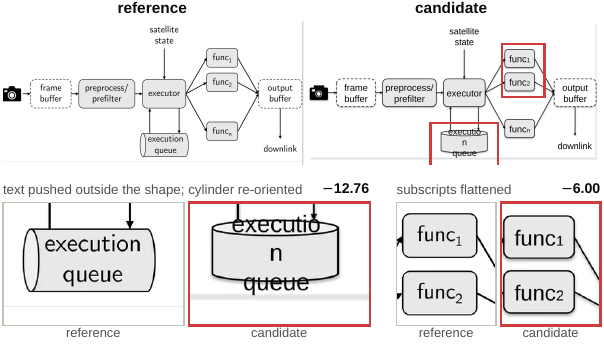}
  \caption{Luna Medium on task~0852 ($S=79.99$). Enlarged regions show text overflow,
  flattened subscripts, and the reoriented cylinder.}
  \label{fig:case_study}
\end{figure}

\begin{figure}[htbp]
  \centering
  \includegraphics[width=0.95\textwidth]{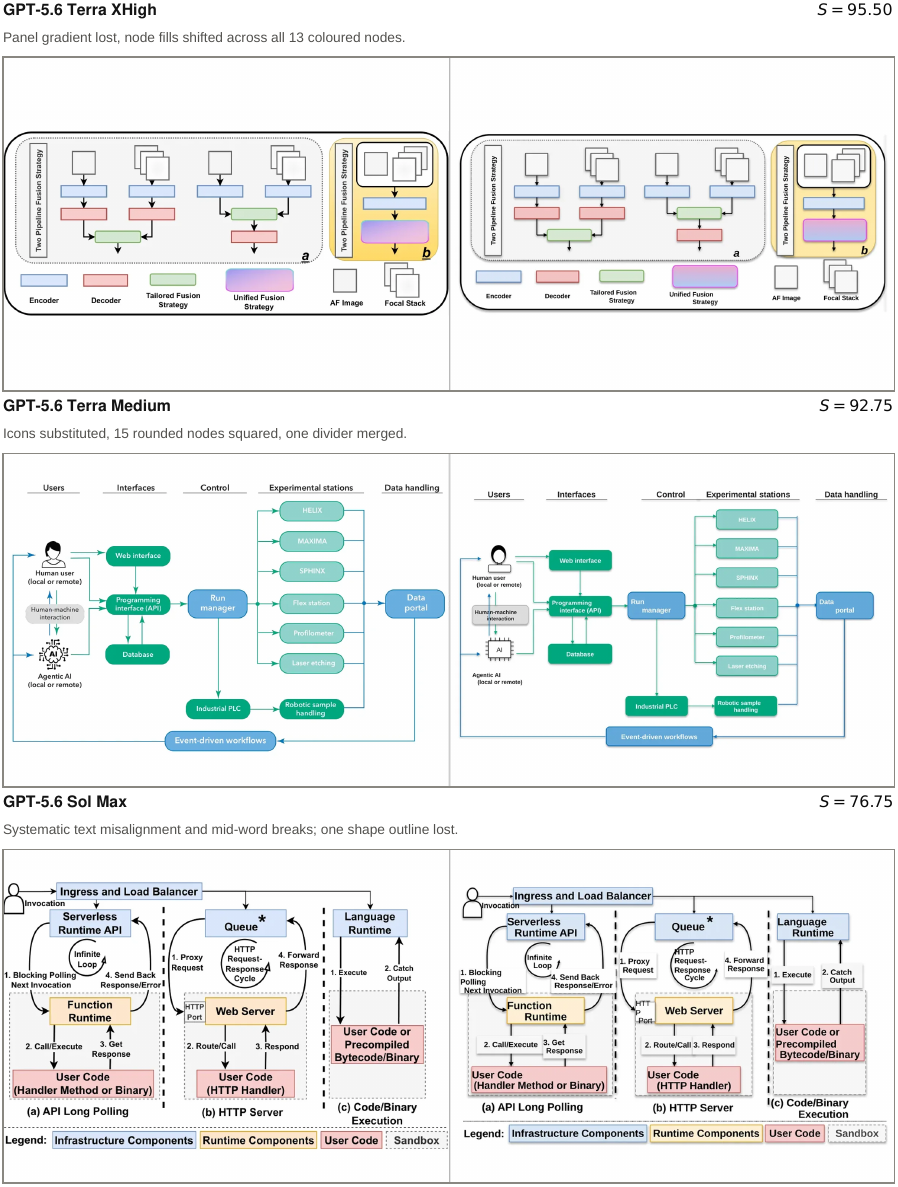}
  \caption{Three additional gate-passing reconstructions with different detail errors. Table~\ref{tab:worked_cases} reports their dimension scores.}
  \label{fig:gallery}
\end{figure}

Table~\ref{tab:worked_cases} decomposes four gate-passing reconstructions.
Figure~\ref{fig:case_study} shows the Luna Medium example; Figure~\ref{fig:gallery} shows the other three.
\begin{table}[H]
  \caption{Detail scores for four gate-passing reconstructions in Figures~\ref{fig:case_study} and~\ref{fig:gallery}. Bold indicates the highest score in each column.}
  \label{tab:worked_cases}
  \centering
  \scriptsize
  \renewcommand{\arraystretch}{1.08}
  \begin{adjustbox}{max width=\linewidth}
  \begin{tabular}{lrrrrp{0.34\linewidth}}
    \toprule
    System & Layout & Text & Graphics & $S$ & Dominant defect \\
    & /30 & /40 & /30 & /100 & \\
    \midrule
    GPT-5.6 Terra XHigh & 28.00 & \textbf{40.00} & 27.50 & \textbf{95.50} & Panel gradient lost; node fills shifted across all 13 coloured nodes \\
    GPT-5.6 Terra Medium & 27.75 & \textbf{40.00} & 25.00 & 92.75 & Icons substituted; 15 rounded nodes squared; one title divider merged \\
    GPT-5.6 Luna Medium & \textbf{28.75} & 23.49 & \textbf{27.75} & 79.99 & One text frame overruns its shape in four ways; cylinder re-oriented \\
    GPT-5.6 Sol Max & \textbf{28.75} & 27.50 & 20.50 & 76.75 & Systematic text misalignment and mid-word breaks; one shape outline missing \\
    \bottomrule
  \end{tabular}
  \end{adjustbox}
\end{table}

\clearpage

\clearpage
\section{Detail scoring and taxonomy}
\label{app:detail_taxonomy}

\paragraph{Structured findings and round aggregation.} Each Stage~3 finding records its
dimension, category, issue subtype, location, description, and affected-instance count. The
judge estimates one affected fraction per category, which is shared by the findings in that
category. Affected counts are retained for inspection but do not directly enter the penalty.
Only rounds that pass both judged gates contribute detail findings. If fewer than two rounds
pass, the candidate receives zero, including when one round fails Stage~1 and another fails
Stage~2. Otherwise, findings are merged by dimension, category, and issue subtype. A finding
reported in any passing round enters the merged set, and repeated findings retain the largest
affected fraction. This counts each issue type once within its category and dimension while
preserving its widest reported extent. The contributing rounds and individual observations
are retained for audit.

\paragraph{Deductions for visual defects.} A finding's deduction depends on its severity and
how much of the diagram it affects. Let $K_r$ contain findings from passing round $r$ and $K_d$
contain merged findings in dimension $d$. For finding $k$, $\rho_{rk}\in[0,1]$ is the category-level
affected fraction assigned to it in round $r$. Its merged fraction is
\begin{equation}
  \rho_k = \max_{r:\,k\in K_r} \rho_{rk}.
  \label{eq:max_ratio}
\end{equation}
The taxonomy assigns a severity base $b_k\in\{3.0,1.5,0.5,0\}$ to high, medium, low, and
diagnostic findings, respectively. The affected fraction scales this base by
\begin{equation}
  m(\rho)=
  \begin{cases}
    1.00, & 0\leq\rho\leq0.02,\\
    1.25, & 0.02<\rho\leq0.10,\\
    1.50, & 0.10<\rho\leq0.25,\\
    2.00, & 0.25<\rho\leq0.50,\\
    2.50, & 0.50<\rho\leq1.
  \end{cases}
  \label{eq:ratio_staircase}
\end{equation}
Using fractions makes an issue affecting four of four nodes more costly than the same issue
affecting four of forty. With dimension caps
$(W_{\mathrm{layout}},W_{\mathrm{text}},W_{\mathrm{graphics}})=(30,40,30)$, the total penalty $P_d$
and retained score $S_d$ are
\begin{equation}
  P_d=\min\!\left(W_d,\sum_{k\in K_d}
  \operatorname{round}_2\!\left[b_k m(\rho_k)\right]\right),\qquad
  S_d=W_d-P_d,
  \label{eq:detail_score}
\end{equation}
where $\operatorname{round}_2$ rounds each deduction to two decimals before summation. These
three retained scores form the visual-quality term in Eq.~\eqref{eq:score}. The deterministic
calculation gives identical scores for identical merged findings and allows rubric weights to
be changed without re-judging the images.

\paragraph{Issue taxonomy.} Table~\ref{tab:detail_taxonomy} lists the issue subtypes and
severity bases used in Eq.~\eqref{eq:detail_score}. Diagnostic entries are recorded without
deductions.
\begingroup
\footnotesize
\setlength{\tabcolsep}{4pt}
\renewcommand{\arraystretch}{1.12}
\begin{longtable}{@{}>{\raggedright\arraybackslash}p{0.17\linewidth}>{\raggedright\arraybackslash}p{0.13\linewidth}>{\raggedright\arraybackslash}p{\dimexpr0.70\linewidth-16pt\relax}@{}}
  \caption{Stage~3 issue taxonomy and severity bases. Diagnostic findings carry zero penalty.}\label{tab:detail_taxonomy}\\
  \toprule
  Category & Severity base & Issue subtype \\
  \midrule
  \endfirsthead
  \multicolumn{3}{l}{Table~\thetable{} (continued)}\\
  \toprule
  Category & Severity base & Issue subtype \\
  \midrule
  \endhead
  \bottomrule
  \endfoot
  \multicolumn{3}{@{}l}{\textbf{Layout and composition}}\\*
  Panel misalignment & High (3.0) & \texttt{missing\_panel}, \texttt{extra\_panel}, \texttt{duplicate\_panel}, \texttt{substituted\_panel}, \texttt{panel\_order}, \texttt{panel\_overlap}, \texttt{panel\_clipping}, \texttt{conspicuous\_blank\_region} \\
  Panel misalignment & Medium (1.5) & \texttt{canvas\_crop}, \texttt{canvas\_aspect\_ratio}, \texttt{canvas\_orientation}, \texttt{panel\_position}, \texttt{panel\_size}, \texttt{panel\_proportion}, \texttt{panel\_spacing}, \texttt{panel\_alignment}, \texttt{separator\_position}, \texttt{other\_panel\_layout} \\
  Panel misalignment & Diagnostic (0.0) & \texttt{global\_scale}, \texttt{global\_centering}, \texttt{outer\_whitespace} \\
  Node layout deviation & High (3.0) & \texttt{missing\_node}, \texttt{extra\_node}, \texttt{duplicate\_node}, \texttt{substituted\_node}, \texttt{reading\_order}, \texttt{grouping}, \texttt{nesting}, \texttt{hierarchy}, \texttt{node\_overlap}, \texttt{node\_occlusion} \\
  Node layout deviation & Medium (1.5) & \texttt{node\_position}, \texttt{node\_spacing}, \texttt{node\_alignment}, \texttt{node\_distribution}, \texttt{grid\_arrangement}, \texttt{padding}, \texttt{z\_order}, \texttt{local\_scale}, \texttt{local\_mirroring}, \texttt{local\_rotation}, \texttt{other\_node\_layout} \\
  Panel style error & Low (0.5) & \texttt{panel\_fill}, \texttt{panel\_border\_color}, \texttt{panel\_border\_width}, \texttt{panel\_border\_pattern}, \texttt{panel\_opacity}, \texttt{panel\_gradient}, \texttt{panel\_texture}, \texttt{panel\_clipping\_mask}, \texttt{panel\_corner\_radius}, \texttt{panel\_glow}, \texttt{panel\_bevel}, \texttt{panel\_outer\_frame}, \texttt{panel\_header\_strip}, \texttt{panel\_title\_tab}, \texttt{panel\_separator\_style}, \texttt{other\_panel\_style} \\
  Panel style error & Diagnostic (0.0) & \texttt{panel\_shadow} \\
  \midrule
  \multicolumn{3}{@{}l}{\textbf{Text and typography}}\\*
  Spelling and omission & High (3.0) & \texttt{missing\_text}, \texttt{extra\_text}, \texttt{substituted\_text}, \texttt{duplicated\_text}, \texttt{reordered\_text}, \texttt{incorrect\_number}, \texttt{incorrect\_identifier}, \texttt{math\_symbol}, \texttt{greek\_letter}, \texttt{subscript}, \texttt{superscript}, \texttt{special\_glyph}, \texttt{wrong\_label\_assignment}, \texttt{other\_text\_content} \\
  Spelling and omission & Low (0.5) & \texttt{capitalization}, \texttt{punctuation}, \texttt{abbreviation}, \texttt{accent}, \texttt{unit} \\
  Text overflow or edge touch & High (3.0) & \texttt{text\_clipping}, \texttt{text\_truncation}, \texttt{outside\_text\_box}, \texttt{text\_text\_overlap}, \texttt{text\_node\_overlap}, \texttt{text\_icon\_overlap}, \texttt{text\_connector\_overlap}, \texttt{text\_arrowhead\_overlap}, \texttt{text\_cell\_overlap}, \texttt{cut\_off\_glyph}, \texttt{other\_text\_fit} \\
  Text overflow or edge touch & Medium (1.5) & \texttt{edge\_touch}, \texttt{insufficient\_text\_padding} \\
  Improper typography & Medium (1.5) & \texttt{font\_family}, \texttt{font\_weight}, \texttt{text\_color}, \texttt{text\_contrast}, \texttt{line\_height}, \texttt{baseline}, \texttt{math\_typesetting}, \texttt{horizontal\_alignment}, \texttt{vertical\_alignment}, \texttt{text\_anchoring}, \texttt{unexpected\_wrap}, \texttt{mid\_word\_break}, \texttt{isolated\_character\_line}, \texttt{line\_count}, \texttt{line\_order}, \texttt{paragraph\_width}, \texttt{text\_rotation}, \texttt{text\_orientation} \\
  Improper typography & Low (0.5) & \texttt{font\_size}, \texttt{font\_style}, \texttt{text\_opacity}, \texttt{letter\_spacing}, \texttt{word\_spacing}, \texttt{kerning}, \texttt{internal\_padding}, \texttt{hyphenation}, \texttt{curved\_text}, \texttt{text\_blur}, \texttt{text\_aliasing}, \texttt{other\_typography} \\
  \midrule
  \multicolumn{3}{@{}l}{\textbf{Local graphics, nodes and connectors}}\\*
  Node shape error & High (3.0) & \texttt{missing\_shape\_primitive}, \texttt{extra\_shape\_primitive} \\
  Node shape error & Medium (1.5) & \texttt{shape\_type}, \texttt{node\_aspect\_ratio}, \texttt{node\_size}, \texttt{node\_rotation}, \texttt{node\_orientation}, \texttt{node\_skew}, \texttt{node\_symmetry}, \texttt{notch\_geometry}, \texttt{folded\_corner}, \texttt{cutout}, \texttt{tab\_geometry}, \texttt{port\_geometry}, \texttt{handle\_geometry}, \texttt{tail\_geometry}, \texttt{pointer\_geometry}, \texttt{other\_node\_geometry} \\
  Node shape error & Low (0.5) & \texttt{node\_corner\_radius} \\
  Node style error & Low (0.5) & \texttt{node\_fill}, \texttt{node\_stroke}, \texttt{node\_border\_width}, \texttt{node\_border\_pattern}, \texttt{node\_color}, \texttt{node\_opacity}, \texttt{node\_transparency}, \texttt{node\_gradient}, \texttt{node\_texture}, \texttt{node\_hatch}, \texttt{node\_highlight}, \texttt{node\_contrast}, \texttt{state\_color}, \texttt{palette\_mapping}, \texttt{three\_dimensional\_treatment}, \texttt{node\_bevel}, \texttt{node\_gloss}, \texttt{internal\_z\_order}, \texttt{style\_consistency}, \texttt{other\_node\_style} \\
  Node shadow error & Diagnostic (0.0) & \texttt{shadow\_presence}, \texttt{shadow\_offset}, \texttt{shadow\_direction}, \texttt{shadow\_blur}, \texttt{shadow\_spread}, \texttt{shadow\_color}, \texttt{shadow\_opacity}, \texttt{glow\_presence}, \texttt{glow\_style}, \texttt{hard\_vs\_soft\_shadow}, \texttt{other\_shadow} \\
  Vector and icon confusion & Medium (1.5) & \texttt{missing\_icon}, \texttt{extra\_icon}, \texttt{substituted\_icon}, \texttt{malformed\_icon}, \texttt{mirrored\_icon}, \texttt{rotated\_icon}, \texttt{distorted\_icon}, \texttt{icon\_state}, \texttt{legend\_symbol}, \texttt{legend\_mapping}, \texttt{rasterized\_vector}, \texttt{grid\_dimensions}, \texttt{grid\_count}, \texttt{primitive\_count}, \texttt{primitive\_order}, \texttt{mini\_diagram\_topology}, \texttt{chart\_axes}, \texttt{chart\_ticks}, \texttt{chart\_legend}, \texttt{chart\_series}, \texttt{chart\_marker}, \texttt{raster\_crop}, \texttt{raster\_stretch}, \texttt{raster\_replacement}, \texttt{other\_local\_graphic} \\
  Vector and icon confusion & Low (0.5) & \texttt{vector\_blur}, \texttt{vector\_pixelation}, \texttt{jagged\_vector}, \texttt{cell\_size}, \texttt{cell\_spacing}, \texttt{cell\_border}, \texttt{cell\_alignment}, \texttt{primitive\_position}, \texttt{primitive\_color}, \texttt{primitive\_fill}, \texttt{primitive\_pattern}, \texttt{raster\_blur} \\
  Line and arrow details & High (3.0) & \texttt{missing\_connector}, \texttt{extra\_connector}, \texttt{wrong\_source\_attachment}, \texttt{wrong\_target\_attachment}, \texttt{wrong\_port}, \texttt{connector\_direction}, \texttt{bidirectionality} \\
  Line and arrow details & Medium (1.5) & \texttt{detached\_endpoint}, \texttt{endpoint\_side}, \texttt{arrowhead\_endpoint}, \texttt{intermediate\_arrowhead}, \texttt{arrowhead\_count}, \texttt{arrowhead\_shape}, \texttt{route\_shape}, \texttt{bezier\_curvature}, \texttt{bend\_count}, \texttt{bend\_location}, \texttt{bend\_radius}, \texttt{loop\_shape}, \texttt{bypass\_shape}, \texttt{junction\_dot}, \texttt{split\_merge\_marker}, \texttt{crossing\_bridge}, \texttt{intersection\_treatment}, \texttt{connector\_overlap\_text}, \texttt{connector\_overlap\_node}, \texttt{connector\_clipping}, \texttt{border\_confusion}, \texttt{other\_connector} \\
  Line and arrow details & Low (0.5) & \texttt{arrowhead\_size}, \texttt{arrowhead\_fill}, \texttt{arrowhead\_outline}, \texttt{arrowhead\_color}, \texttt{arrowhead\_angle}, \texttt{parallel\_spacing}, \texttt{line\_pattern}, \texttt{dash\_sequence}, \texttt{dash\_gap}, \texttt{line\_width}, \texttt{line\_color}, \texttt{line\_opacity}, \texttt{line\_cap}, \texttt{line\_join}, \texttt{line\_blur}, \texttt{line\_glow} \\
  Line and arrow details & Diagnostic (0.0) & \texttt{line\_shadow} \\
\end{longtable}
\endgroup

\end{document}